\documentclass{article}

\usepackage[preprint]{neurips_2025}

\usepackage[utf8]{inputenc} 
\usepackage[T1]{fontenc}    
\usepackage{hyperref}       
\usepackage{url}            
\usepackage{booktabs}       
\usepackage{amsfonts}       
\usepackage{nicefrac}       
\usepackage{microtype}      

\usepackage{graphicx}
\usepackage{wrapfig}
\usepackage{subcaption}
\usepackage{multirow}

\usepackage{amsmath}
\usepackage{amsthm}
\usepackage{mathtools}
\usepackage{mathrsfs}
\usepackage{algorithm}
\usepackage{algpseudocode}

\usepackage{amsmath}
\usepackage{amssymb}

\usepackage{verbatim}
\usepackage{pifont}
\usepackage{bm}

\usepackage{colortbl}
\usepackage[dvipsnames, svgnames, x11names]{xcolor}
\definecolor{aliceblue}{rgb}{0.94, 0.97, 1.0}
\definecolor{deeppink}{RGB}{255,20,147}
\definecolor{mygray}{gray}{0.9}

\definecolor{myred}{rgb}{0.90, 0.43, 0.31}
\definecolor{myyellow}{rgb}{0.91, 0.76, 0.41}
\definecolor{mygreen}{rgb}{0.16, 0.61, 0.56}
\definecolor{myblue}{rgb}{0.29, 0.51, 0.82}
\definecolor{myblue2}{rgb}{0.53, 0.70, 0.92}

\usepackage{amsmath,amsfonts,bm}

\def\eqref#1{equation~\ref{#1}}

\def\1{\bm{1}}

\def\mB{{\bm{B}}}

\def\mD{{\bm{D}}}

\def\mF{{\bm{F}}}

\def\mI{{\bm{I}}}

\def\mP{{\bm{P}}}

\def\mV{{\bm{V}}}

\def\mX{{\bm{X}}}

\DeclareMathAlphabet{\mathsfit}{\encodingdefault}{\sfdefault}{m}{sl}
\SetMathAlphabet{\mathsfit}{bold}{\encodingdefault}{\sfdefault}{bx}{n}

\newcommand{\R}{\mathbb{R}}

\title{Uni3D-MoE: Scalable Multimodal 3D Scene Understanding via Mixture of Experts}

\author{
  Yue Zhang  \\
  Zhejiang University \\
  \And
  Yingzhao Jian \\
  Zhejiang University \\
  \And
  Hehe Fan \\
  Zhejiang University \\ 
  \AND
  Yi Yang \\
  Zhejiang University \\
  \And
  Roger Zimmermann \\
  National University of Singapore \\ 
}

\begin{document}

\maketitle

\begin{abstract}
Recent advancements in multimodal large language models (MLLMs) have demonstrated considerable potential for comprehensive 3D scene understanding. 
However, existing approaches typically utilize only one or a limited subset of 3D modalities, resulting in incomplete representations of 3D scenes and reduced interpretive accuracy. Furthermore, different types of queries inherently depend on distinct modalities, indicating that uniform processing of all modality tokens may fail to effectively capture query-specific context. 
To address these challenges, we propose Uni3D-MoE, a sparse Mixture-of-Experts (MoE)-based 3D MLLM designed to enable adaptive 3D multimodal fusion. Specifically, Uni3D-MoE integrates a comprehensive set of 3D  modalities, including multi-view RGB and depth images, bird’s-eye-view (BEV) maps, point clouds, and voxel representations. At its core, our framework employs a learnable routing mechanism within the sparse MoE-based large language model, dynamically selecting appropriate experts at the token level. Each expert specializes in processing multimodal tokens based on learned modality preferences, thus facilitating flexible collaboration tailored to diverse task-specific requirements.
Extensive evaluations on standard 3D scene understanding benchmarks and specialized datasets demonstrate the efficacy of Uni3D-MoE.
\end{abstract}

\section{Introduction}
3D scene understanding is fundamental for intelligent systems such as robotic navigation~\citep{werby2024hierarchical,yin2024sg,gu2024conceptgraphs} and autonomous driving~\citep{kong2025multi,han2025dme,zhou2024drivinggaussian,wei2024editable}. Recent advances in multimodal large language models (MLLMs) have demonstrated considerable potential for enhancing the interpretation and analysis of complex 3D environments~\citep{zha2025enable,hu2024scenecraft,fu2024scene,chen2024grounded,xu2024comp4d,yang2024llm}.
Usually, existing methods for multimodal 3D scene understanding leverage specific combinations of input modalities. 
For instance, Chat-3D~\citep{wang2023chat} constructs 3D MLLMs primarily from point clouds. Chat-Scene~\citep{huang2024chat} combines multi-view RGB images and point cloud data. GPT4Scene~\citep{qi2025gpt4scene} integrates RGB images of multiple views with bird's eye view (BEV) representations.  
Video-3D LLM~\citep{zheng2024video} leverages positional video and 3D coordinates to generate spatially-aware representations.

\begin{wrapfigure}{r}{0.53\linewidth}  
    \centering
    \vspace{-1.5em}
    \includegraphics[width=0.98\linewidth]{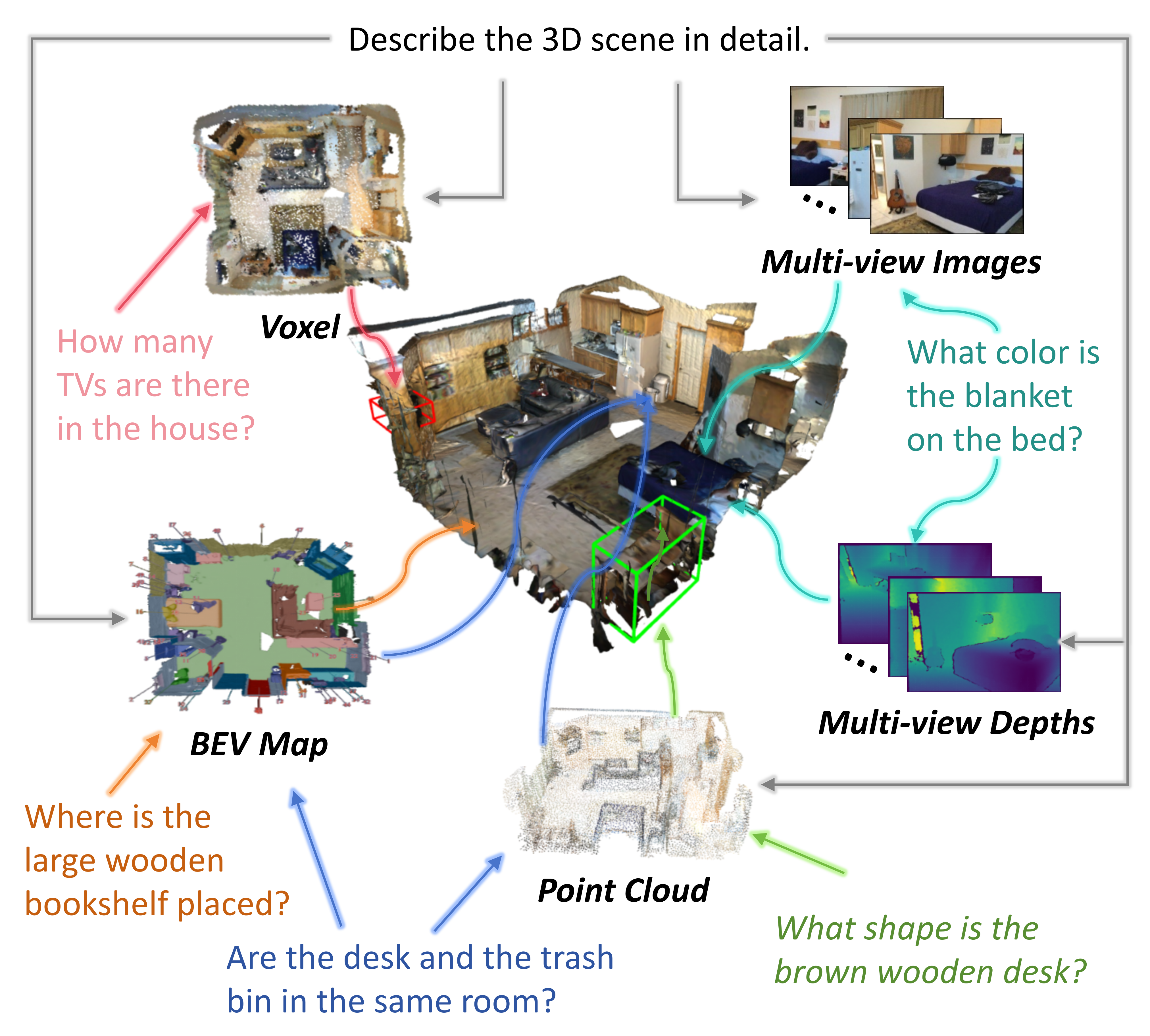}
    \vspace{-0.5em}
    \caption{Challenges in 3D scene understanding. (1) Limited modalities may not provide enough scene information. (2) Different question types have varying dependencies on modalities. Existing methods typically treat all modality tokens equally, without adapting to question-specific modality preferences.} 
    \label{fig:teaser}
    \vspace{-1.2em}
\end{wrapfigure}

Despite these advancements, leveraging diverse modalities effectively for comprehensive 3D understanding remains challenging, as shown in Fig.~\ref{fig:teaser}. Two critical limitations persist: 1) Existing methods usually rely on a limited subset of available modalities, potentially omitting essential information due to occlusions or viewpoint restrictions. For example, relying solely on multi-view RGB images might fail to capture obscured objects, complicating queries such as ``How many TVs are in the house?'' 
2) Different question types exhibit varying dependencies on specific modalities. Queries about object geometry, like ``What shape is the wooden desk?'', are better addressed with geometric modalities such as point clouds, while color-based queries (e.g., ``What color is the blanket on the bed?'') primarily use visual cues from RGB or depth images. Current dense architectures typically process all modalities uniformly, hindering adaptive alignment with query-specific modality preferences.

In this paper, we propose \textbf{Uni3D-MoE}, a scalable and adaptive multimodal 3D scene understanding framework built upon a sparse Mixture-of-Experts (MoE) architecture. Uni3D-MoE comprises three key components: 1) modality-specific encoders that extract features from multiple input modalities, including multi-view RGB and depth images, BEV maps, point clouds, and voxels; 2) modality-alignment adapters that unify these diverse modality-specific features into a common latent representation; and 3) sparse MoE modules integrated within the large language model (LLM), featuring a learnable routing mechanism to dynamically select appropriate experts for each modality token.
Our Uni3D-MoE has two merits: 1) by integrating comprehensive 3D modalities, the model achieves more complete and accurate scene representations; 2) the routing mechanism selectively activates relevant expert pathways based on modality tokens, enabling specialized, adaptive processing tailored to each modality.

Extensive experiments on public 3D scene understanding benchmarks~\citep{azuma2022scanqa, chen2021scan2cap, ma2022sqa3d} and datasets curated around specific question types demonstrate the effectiveness of Uni3D-MoE.
In particular, our model achieves CIDEr gains of 46.9 on the location task and 51.9 on the color task through comprehensive modality details, with further improvements of 6.6 and 10.2 from introducing MoE. The main contributions are summarized as follows:
\begin{itemize}
    \item \textbf{Unified 3D MoE architecture.} We propose Uni3D-MoE, the first unified sparse MoE-based MLLM explicitly designed for 3D scene understanding, supporting a wide range of modalities, including multi-view RGB-D images, BEV maps, point clouds, and voxels. 

    \item \textbf{Exploring MoE for Adaptive 3D Modality Fusion.}
    As the first time, we employ sparse MoE to adaptively fuse 3D modalities.
    The adaptive routing effectively enhances multimodal fusion tailored to specific queries.

    \item 
    \textbf{Enhanced performance. } Extensive empirical evaluation demonstrates that Uni3D-MoE significantly outperforms existing methods on several 3D scene understanding tasks. 
  
\end{itemize}

\section{Related Work}
\textbf{3D Vision-language Learning.}
Early research mainly leveraged point clouds~\citep{xu2024pointllm, guo2023point, liu2024uni3d, cao2025objvariantensemble, qi2024gpt4point} or voxels~\citep{fu2024scene, zhu2024unifying, yang2025lidar} to model objects and scenes, enabling LLMs to perform basic 3D grounding~\citep{yang2024llm, chen2024grounded, fei2024kestrel, wang2025liba} and question answering~\citep{zhu20254d, szymanska2024space3d, li2025embodied}.
Motivated by the spatial ability of self-supervised 2D encoders, subsequent works explore projecting 2D features into 3D space~\citep{zhu2024llava, yang2025lidar}, which enhances the comprehension of fine-grained details and complex structures.
Recently, 2D videos~\citep{fu2024scene, xiong20253ur, qi2025gpt4scene, zheng2024video} and BEV~\citep{qi2025gpt4scene} have also demonstrated comparable performance to 3D representations, prompting a renewed focus on 2D information in spatial understanding.
While these efforts have significantly advanced the field~\citep{jia2024sceneverse, wang2023image, li2024uni3dl, zhang2024vision, ma2024llms}, most of them focus on exploiting the strengths of a single modality. The exploration of complementary advantages across multiple modalities remains an open and compelling research topic.

\textbf{Mixture of Experts.}
The Mixture-of-Experts (MoE) framework achieves comparable performance to dense models while activating far fewer parameters during inference~\citep{zhou2022mixture, cai2024survey, du2024revisiting, song2024promoe, xue2024openmoe}.
MoE is typically categorized into two types based on whether the routing is learned: hard and soft routers.
The hard router~\citep{DBLP:conf/nips/BaoW0LMASPW22, shen2023scaling} is suitable for scenarios with clear modality boundaries and predefined expert assignments.
It directly routes different modalities (e.g., images, text) to designated experts~\citep{wang2023chat, wang2023image}.
However, hard routing is inflexible, unable to capture token-level semantics or adapt expert selection to tasks~\citep{DBLP:journals/corr/abs-2401-15947}.
To improve model adaptability, recent dense LLMs have introduced sparse soft routing mechanisms~\citep{yue2024ada, fedus2022switch}, such as LLaMA-MoE~\citep{zhu2024llama}, MoE-LLaVA~\citep{DBLP:journals/corr/abs-2401-15947}, and Uni-MoE~\citep{DBLP:journals/pami/LiJHWZLMZ25}.
These models primarily operate on 1D text and 2D visual inputs.
3D-MoE~\citep{ma20253d} and MiniGPT-3D~\citep{tang2024minigpt} initially explore MoE for 3D tasks but lack unified modeling of diverse 3D modalities like voxels, BEV and multi-view images for scene understanding.
To this end, we aim to develop a 3D MLLM with soft-routing MoE to achieve unified and adaptive fusion of diverse 3D modalities.

\begin{figure}[t!]
    \centering
    \includegraphics[width=1.0\linewidth]{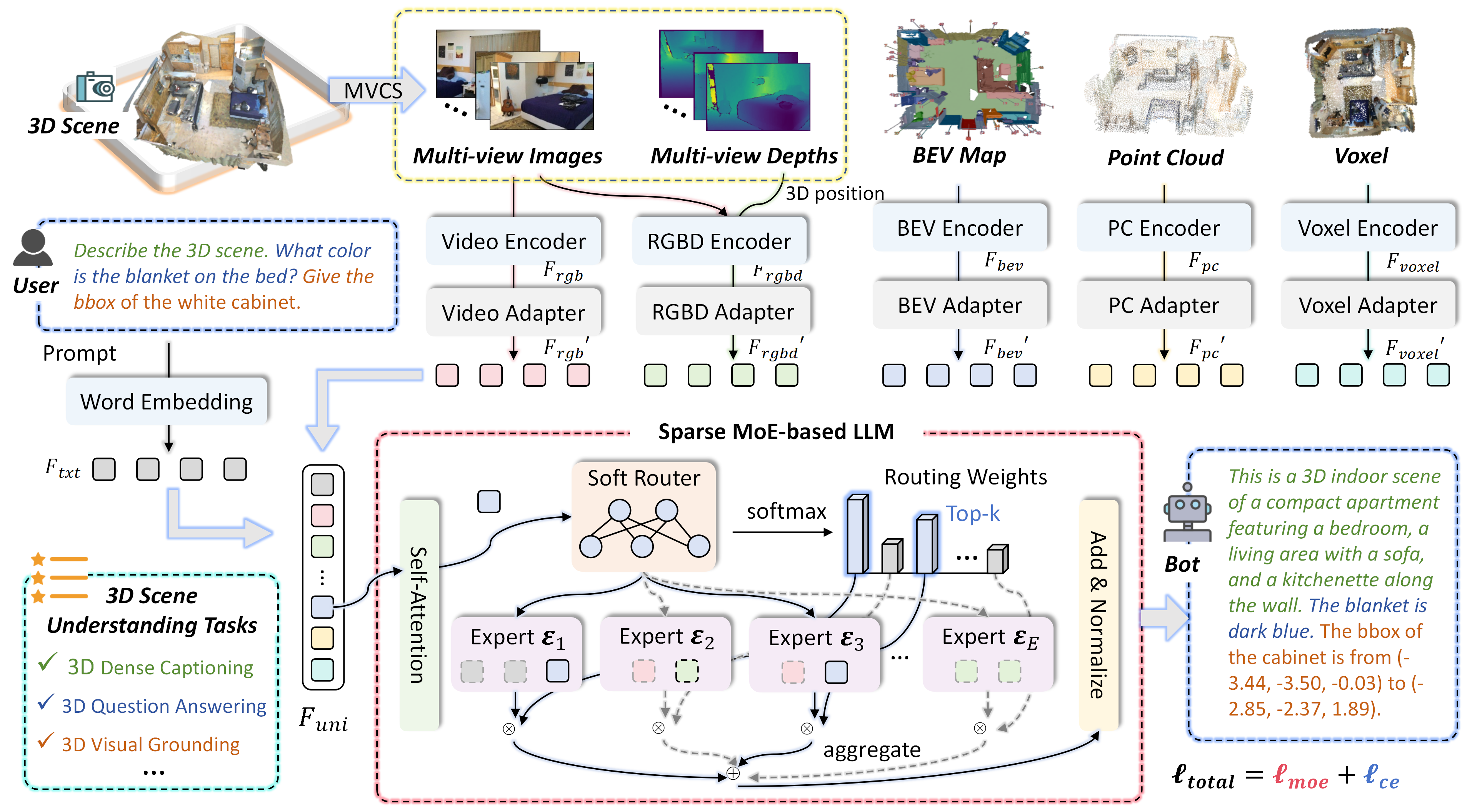}
    \caption{Overview of our method. Uni3D-MoE covers major input modalities of 3D scenes, including RGB/depth images, BEV maps, point clouds, and voxels. To ensure informative spatial coverage, multi-view images are selected using a Maximum Voxel Coverage Sampling (MVCS) algorithm.
    Each modality is encoded by a modality-specific encoder and aligned via lightweight adapters. 
    The resulting 3D visual tokens, together with text tokens, are then fed into a sparse Mixture-of-Expert (MoE)-based LLM.
    A learnable soft router dynamically assigns each token to a subset of suitable experts for specialized processing.
    The model is optimized with a joint objective combining cross-entropy loss $l_{ce}$ and a sparsity-aware expert balancing loss $l_{moe}$.} 
    \label{fig:overview}
    \vspace{-1em}
\end{figure}

\section{Method}
\textbf{Overview.}
Uni3D-MoE is a unified 3D MLLM framework that leverages sparse MoE for adaptive scene understanding.
Fig.~\ref{fig:overview} illustrates the architecture of Uni3D-MoE, which contains 3D scene feature encoders, feature alignment adapters, and a sparse MoE-enhanced LLM.
First, we introduce the 3D scene feature extractor, designed to handle diverse 3D modalities and produce unified feature tokens (Sec.~\ref{sec:extractor}).
Then, we detail the learnable soft router, which selectively activates expert pathways to enable token-level specialization (Sec.~\ref{sec:moe}).
Finally, we present the training strategy~\ref{sec:stg} and optimization objectives~\ref{sec:obj}.

\subsection{3D Scene Feature Extractor}~\label{sec:extractor}
\noindent
\textbf{Modality Data Preparation.}
First, we employ the Maximum Voxel Coverage Sampling (MVCS) algorithm to select informative keyframes.
Compared with previous approaches~\citep{zheng2024video}, our improved MVCS achieves 100× speed-up in computing coverage by using camera poses instead of depth images. Additionally, we enhance frame quality through voxel weighting, depth pruning, and blur image filtering. Further algorithmic details are provided in the Appendix. 
Then, to provide global spatial context, we render BEV maps with explicit semantic segmentation cues.

\textbf{Modality-specific Feature Extraction.}
Uni3D-MoE employs modality-specific encoders to capture more comprehensive representations of 3D scenes.
Specifically, multi-view RGB images are encoded using a pre-trained DINOv2~\citep{oquab2023dinov2} to obtain $\mF_{rgb}$.
For multi-view RGB-D inputs, we first extract 2D patches via CLIP~\citep{radford2021learning}, and then integrate corresponding 3D spatial positions derived from depth maps, generating spatially-aware RGB-D features $\mF_{rgbd}$.
We also use DINOv2~\citep{oquab2023dinov2} to extract BEV features $\mF_{bev}$.
Point clouds, downsampled via Farthest Point Sampling (FPS)~\citep{qi2017pointnet++}, are processed through PointNet++~\citep{qi2017pointnet++}, yielding $\mF_{pc}$.
Voxel grids are voxelized and hierarchically encoded by Mask3D's~\citep{schult2023mask3d} sparse convolutional U-Net to produce $\mF_{voxel}$.

\textbf{Modality Feature Alignment.}
Subsequently, tokens from five modalities are aligned to the text space via respective adapters: 
$\mF_{m}^{\prime} = \text{Adapter}_{m}(\mF_{m}) \in \R^{N_{m} \times D_{txt}}$, where $m \in \{\text{rgb}, \text{rgbd}, \text{bev}, \text{pc}, \text{voxel}\}$, $N_{m}$ is the token count of modality $m$ and $D_{txt}$ is the target embedding dimension.
Finally, the text prompt feature $\mF_{txt}$, combined with modality-aligned features $\mF_{m}^{\prime}$, composes the unified 3D scene representation:
$\mathcal{F}_{uni} = \{ \mF_{txt}, \mF_{rgb}^{\prime}, \mF_{rgbd}^{\prime}, \mF_{bev}^{\prime}, \mF_{pc}^{\prime}, \mF_{voxel}^{\prime}\} \in \R^{N_{uni} \times D_{txt}}$, where $N_{uni} = \sum_m N_m$ denotes the total number of multimodal tokens.

\subsection{Soft Routing for Expert Selection}\label{sec:moe}
The MoE module employs a learnable soft routing mechanism to achieve intelligent token-to-expert assignment.
Given a token $f_i \in \mathcal{F}_{uni}$ and a set of $E$ experts $\{\mathcal{E}_1, \mathcal{E}_2, \ldots, \mathcal{E}_E\}$, a lightweight routing network computes an affinity score $s_i^{(e)}$ between $f_i$ and each expert $\mathcal{E}_e$, where $e \in \{1, 2, \ldots, E\}$.
These scores are then normalized into a probability distribution:
\begin{equation}\label{eq:routing}
\pi_i^{(e)} = \frac{\exp(s_i^{(e)})}{\sum_{j=1}^{E} \exp(s_i^{(j)})}, \quad \text{where} \quad s_i^{(e)} = \bm{w}_e^\top f_i.
\end{equation}
Here, $\mathbf{w}_e \in \mathbb{R}^D$ denotes the expert-specific routing parameter for expert $\mathcal{E}_e$, and $\pi_i^{(e)}$ represents the routing probability of token $f_i$ to expert $\mathcal{E}_e$.
Each token $f_i$ is routed to its top-$k$ experts with the highest probabilities, and the corresponding outputs are aggregated as:
\begin{equation}\label{eq:moe_output}
\hat{f}_i = \sum_{e \in \mathcal{S}_i} \pi_i^{(e)} \cdot \mathcal{E}_e(f_i),
\end{equation}
where $\mathcal{S}_i \subseteq \{1, 2, \ldots, E\}$ is the set of top-$k$ selected experts for $f_i$, and $\mathcal{E}_e(f_i)$ is the output of expert $\mathcal{E}_e$ applied to token $f_i$.

To balance expert utilization and ensure specialized routing, we incorporate sparsity-aware expert balancing loss, detailed in Sec.~\ref{sec:obj}.
In this way, Uni3D-MoE achieves adaptive multimodal fusion within each expert, thus accommodating prompt-specific requirements.

\subsection{Training Strategy}\label{sec:stg}
\begin{wrapfigure}{r}{0.5\linewidth}  
    \centering
    \vspace{-1.5em}
    \includegraphics[width=\linewidth]{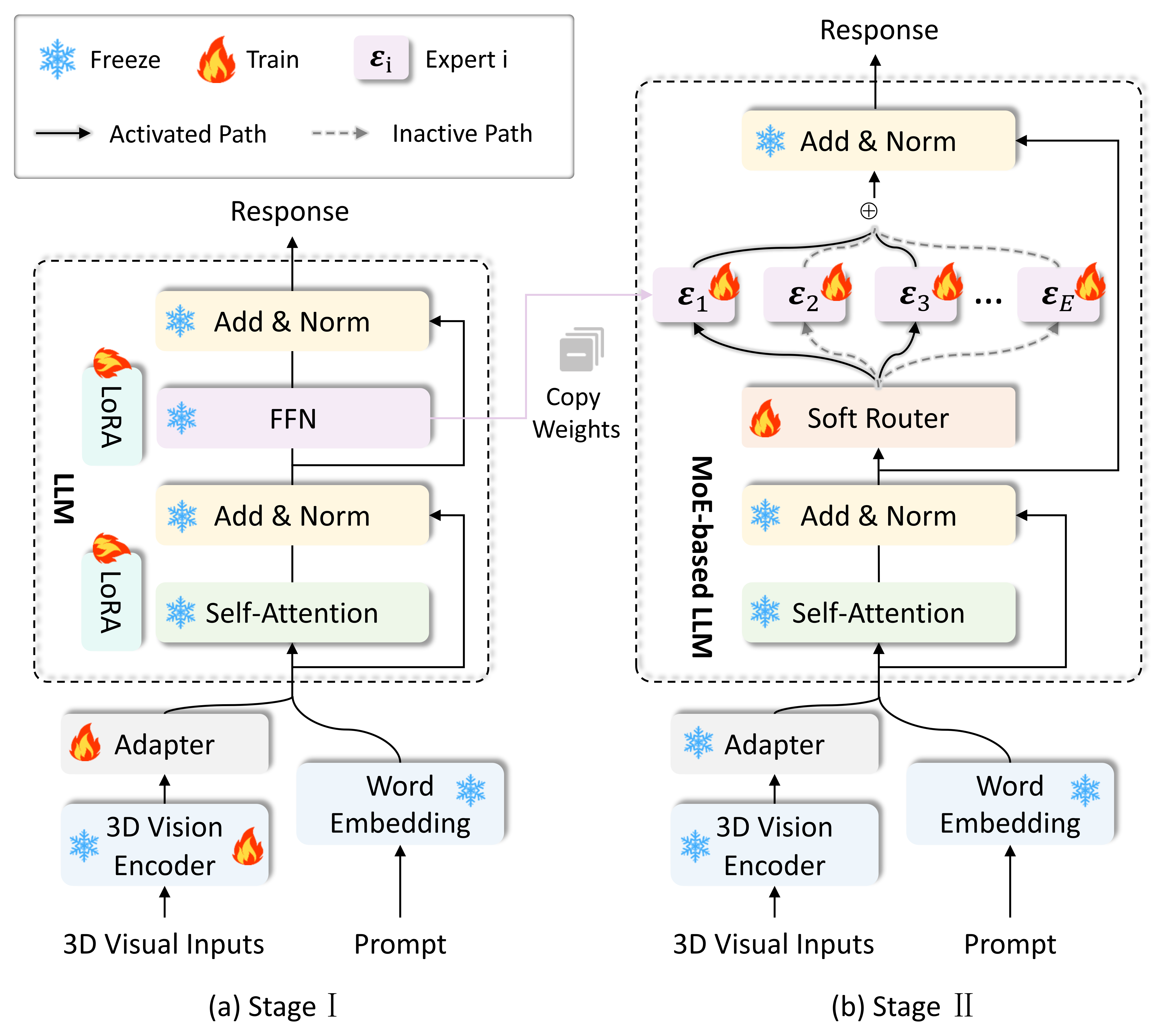}
    \caption{Overview of two-stage training strategy.} 
    \vspace{-2em}
    \label{fig:train-stage}
\end{wrapfigure}
As shown in Fig.~\ref{fig:train-stage}, we design a progressive two-stage training strategy for Uni3D-MoE.

\textbf{Stage \uppercase\expandafter{\romannumeral1}:}
The goal of this stage is to align 3D visual representations with the textual space, enabling the LLM to capture semantic cues from diverse modalities. 
During this stage, the modality-specific adapters and LoRA-injected layers within the LLM are jointly trained, while all visual encoders remain frozen except for the spatial-aware RGBD module and the point cloud encoder.
The model is trained with complex instructions spanning multiple downstream tasks.
We avoid introducing MoE at this stage due to optimization instability when replacing the dense LLM directly. Instead, we refine the model's instruction-following and generation capabilities to prepare for sparse training.

\textbf{Stage \uppercase\expandafter{\romannumeral2}:}
The objective of this stage is to incorporate the sparse MoE architecture to expand model capacity and enable expert specialization.
Inspired by~\citep{DBLP:journals/corr/abs-2401-15947}, we replicate the feed-forward network (FFN) multiple times to initialize expert modules. 
At this stage, only the soft router and expert modules are trainable, while all other parameters are kept frozen.
To guide the routing process and promote expert diversity, we introduce a sparsity-aware expert balancing loss $l_{moe}$ alongside the standard cross-entropy loss $l_{ce}$.
The training data remains the same as in Stage \uppercase\expandafter{\romannumeral1}, ensuring continuity and stability during the transition to sparse expert routing.

\subsection{Training Objective}\label{sec:obj}
All tasks are standardized into a unified user-assistant interaction format.
During training stage \uppercase\expandafter{\romannumeral1}, the training objective is to minimize the autoregressive cross-entropy loss on the generated text:
\begin{equation}\label{eq:l_ce}
    \mathcal{L}_{\text{ce}} = - \sum_{t=1}^{T} \log P_\theta(\mathcal{Y}_t \mid \mathcal{Y}_{<t}, \mathcal{F}_{uni}), 
\end{equation}
where $\mathcal{Y}_t$ is the $t$-th target token, $\mathcal{Y}_{<t}$ denotes previously generated tokens, $\mathcal{F}_{uni}$ is the unified multimodal context, and $\theta$ represents trainable parameters.

In Stage \uppercase\expandafter{\romannumeral2}, we introduce a sparse MoE mechanism and incorporate sparsity-aware expert balancing loss to encourage expert diversity~\citep{fedus2022switch}:
\begin{equation}\label{eq:l_moe}
\begin{array}{c}
\mathcal{L}_{\text{moe}} =  E \cdot \displaystyle \sum_{e=1}^{E} \hat{p}^{(e)} \cdot \bar{\pi}^{(e)}, \\
    \hat{p}^{(e)} = \displaystyle \dfrac{1}{N_{uni}} \sum_{i=1}^{N_{uni}} \mathbf{1} \left\{ \arg\max_{j} \pi_i^{(j)} = e \right\}, \quad
    \bar{\pi}^{(e)} = \dfrac{1}{N_{uni}} \sum_{i=1}^{N_{uni}} \pi_i^{(e)}, 
\end{array}
\end{equation}
where $\hat{p}^{(e)}$ denotes the fraction of tokens routed to expert $e$, and $\bar{\pi}^{(e)}$ is the average routing probability to expert $\mathcal{E}_e$. 
Consistent with the above, $E$ is the number of experts, $N_{uni}$ is the total number of unified tokens, and $\pi_i^{(e)}$ is the routing probability from token $i$ to expert $\mathcal{E}_e$.  
Here, $\mathbf{1}\{\cdot\}$ denotes the indicator function, which returns 1 if the condition holds and 0 otherwise.

The final training objective is defined as the sum of $\mathcal{L}_{\text{ce}}$ and $\mathcal{L}_{\text{moe}}$, where $\lambda$ is a balancing coefficient:
\begin{equation}\label{eq:l_total}
    \mathcal{L}_{\text{total}} =  \mathcal{L}_{\text{ce}} + \lambda
 \cdot \mathcal{L}_{\text{moe}}. 
\end{equation}

\section{Experiments}
\subsection{Experiment Settings}\label{sec:exper-set}
\textbf{Datasets.}
We construct a unified training corpus by aggregating multiple 3D scene understanding datasets built on ScanNet~\citep{dai2017scannet}, which contains $1,513$ indoor RGB-D scans with extensive 2D and 3D annotations. 
The training data covers dense captioning (Scan2Cap~\citep{chen2021scan2cap}), visual question answering (ScanQA\citep{azuma2022scanqa}, SQA3D~\citep{ma2022sqa3d}), and single- and multi-object visual grounding (ScanRefer~\citep{chen2020scanrefer}, Multi3DRefer~\citep{zhang2023multi3drefer}). 
All data are reformatted into a unified user–assistant interaction format.
During evaluation, besides standard 3D scene understanding benchmarks~\citep{azuma2022scanqa, chen2021scan2cap, ma2022sqa3d}, we also use datasets curated around specific question types.
Additional details are provided in the Appendix.

\textbf{Model Details.}
We initialize our LLM from LLaVA-v1.5-7B~\citep{liu2023visual} and introduce MoE module into layers 8, 12, 16, 20, 24, and 28 at the second training stage. 
Each MoE layer comprises 8 experts, with the top-2 experts selected for each token at inference time. Following~\citep{DBLP:journals/corr/abs-2401-15947, DBLP:journals/pami/LiJHWZLMZ25}, a load-balancing coefficient of $\alpha = 0.01$ is applied to promote expert utilization diversity.

\textbf{Training Details.}
We employ a two-stage training strategy: 2 epochs in stage \uppercase\expandafter{\romannumeral1} and 1 epoch in stage \uppercase\expandafter{\romannumeral2}, both with batch size 8.
Both stages utilize the AdamW optimizer with a constant learning rate of 2e-5. 
A warm-up schedule with a warmup ratio of 0.03 followed by cosine decay is applied independently to each stage.
The input sequence length is capped at 4096 tokens. 
To optimize training efficiency and memory usage, we use BF16-based mixed-precision training and leverage DeepSpeed ZeRO-2 offloading.

\textbf{Evaluation Metrics.}
Following~\citep{zhu2024llava, huang2024chat, wang2025ross3d}, we adhere to the commonly used metrics to comprehensively evaluate our method across multiple tasks.
Specifically, for ScanQA~\citep{azuma2022scanqa}, we evaluate the top-1 predicted answers using the exact match accuracy (EM@1), the refined exact match protocol (EM-R@1), F1 score, BLEU-1~\citep{papineni2002bleu}, BLEU-4~\citep{papineni2002bleu}, METEOR~\citep{banerjee2005meteor}, ROUGE~\citep{lin2004rouge}, and CIDEr~\citep{vedantam2015cider}.
For SQA3D~\citep{ma2022sqa3d}, we use EM@1.
For Scan2Cap~\citep{chen2021scan2cap}, we combine CIDEr with an IoU threshold of 0.5 between predicted and reference bounding boxes.

\subsection{Comparison with State-of-the-art Methods} %
\textbf{Comparison Results.}
Table~\ref{tab:results-scanqa} presents the comparative evaluation results on downstream 3D tasks, including ScanQA~\citep{azuma2022scanqa}, SQA3D~\citep{ma2022sqa3d}, and Scan2Cap~\citep{chen2021scan2cap} benchmarks.
Uni3D-MoE exhibits superior performance on the ScanQA benchmark~\citep{chen2021scan2cap}, surpassing existing state-of-the-art methods across multiple metrics.
Specifically, it achieves relative improvements of $11.7\%$ on EM@1,  $16.8\%$ on BLEU-4, and $6.0\%$ on CIDEr.
Furthermore, Uni3D-MoE outperforms LLAVA-3D~\citep{zhu2024llava} by $2.7\%$ on EM@1 on SQA3D benchmark~\citep{ma2022sqa3d}.
On the Scan2Cap~\citep{chen2021scan2cap} benchmark, Uni3D-MoE achieves an improvement of $5.8\%$ on CIDEr@0.5 compared to the previous advanced method PQ3D~\citep{zhu2024unifying}.
More comprehensive results, including visual grounding evaluations, are provided in the Appendix.

\textbf{Analysis.}
The performance benefits from the synergistic effect of heterogeneous modalities, where each modality contributes complementary and modality-specific cues.
Furthermore, sparse expert routing further enhances this by adaptively selecting the most relevant modalities for each input, leading to more precise 3D scene understanding.

\begin{table}[t!]
    \centering
    \caption{Evaluation results of 3D question answering on ScanQA~\citep{azuma2022scanqa} and SQA3D~\citep{ma2022sqa3d}, as well as 3D dense caption on Scan2Cap~\citep{chen2021scan2cap}. EM@1 refers to the top-1 exact match accuracy; BLEU-1, BLEU-4, METEOR, and CIDEr denote text similarity scores between the predicted and ground-truth answer. For Scan2Cap~\citep{chen2021scan2cap}, CIDEr is reported at IoU threshold of 0.5. $\star$ indicates that high-resolution settings are not used. We highlight the best performance in {\color{myred}\textbf{red}} and the second-best in {\color{myblue}\textbf{blue}}.}
    \resizebox{\textwidth}{!}{
    \begin{tabular}{ccccccccc} 
    \toprule
     \multirow{2}{*}{\textbf{Method}}  & \multicolumn{6}{c}{\textbf{ScanQA}} & \textbf{SQA3D} & \textbf{Scan2Cap} \\
     \cmidrule(lr){2-7} \cmidrule(lr){8-8}  \cmidrule(lr){9-9} 
      & EM@1 & BLEU-1 & BLEU-4 & METEOR & ROUGE & CIDEr & EM@1 & CIDEr\\
     \midrule
     \textbf{\textit{Task-specific}} \\
      ScanQA~\citep{azuma2022scanqa} &  21.1  & 30.2 &  10.1 & 13.1 &  33.3 & 64.9 & 47.2 & -\\
      3D-VLP ~\citep{jin2023context} & -  & 30.5 & 11.2 & 13.5 & 34.5 & - & 54.9 &  55.0 \\ 
      3D-VisTA~\citep{zhu20233d} & 22.4  & - & - & 13.9 &  35.7 & - & 48.5 & 66.9 \\

      \midrule
      \textbf{\textit{2D LLMs}} \\
      InternVL2-8B~\citep{lu2025internvl} & -  & - & 3.3 & 14.5 & 34.3 & 62.5 & 33.0 & -\\
      Qwen2-VL-7B~\citep{wang2024qwen2} & -  & 27.8 & 3.0 & 11.4 & 29.3 & 53.9 & 40.7 &  -\\
      LLaVA-Video~\citep{zhang2024video} & - & - & 3.1 & 17.7 & 44.6 & 88.7 & 48.5 & -\\ 
      
      \midrule
      \textbf{\textit{3D LLMs}}\\
      PQ3D~\citep{zhu2024unifying} & 20.0  & 36.1 & - & 13.9 & - & 65.2 & 47.1 & {\color{myblue}\textbf{80.3}} \\
      LAMM~\citep{yin2023lamm} & - & - & 5.8 & -& - & 42.4 & - & -\\
      3D-LLM~\citep{hong20233d} & 20.5 & 39.3 & 12.0 & 14.5 & 35.7 & 69.4 & - & - \\
      Chat-3D~\citep{wang2023chat} & -  &29.1 & 6.4 & 11.9 & 28.5 &  53.2 & - & -\\
      Chat-3D V2~\citep{DBLP:journals/corr/abs-2312-08168} & 22.9 & 38.4 & 7.3 & 16.1 & 40.1 & 77.1 & 54.7 & - \\
      Chat-Scene~\citep{huang2024chat} & 21.6 & 43.2 & 14.3 & 18.0 & 41.6 & 87.7 & 54.6 & 77.1\\
      LL3DA~\citep{chen2024ll3da} & - & - & 13.5 & 15.9 & 37.3 & 76.8 & - & 65.2\\
     LLaVA-3D~\citep{zhu2024llava} & 27.0 & - & 14.5 & {\color{myred}\textbf{20.7}} & {\color{myred}\textbf{50.1}} & {\color{myblue}\textbf{91.7}} & {\color{myblue}\textbf{55.6}}  & 79.2 \\
      LEO~\citep{huang2024embodied} & -  & - & 11.5 & 16.2 & 39.3 & 80.0 & 50.0 & 72.4\\
      Scene-LLM~\citep{fu2024scene} & {\color{myblue}\textbf{27.2}}  & - & 12.0 & 16.6 & 40.0 & 80.0 & 54.2 & 37.9\\
      GPT4Scene$^{\star}$ ~\citep{qi2025gpt4scene} &  - & {\color{myblue}\textbf{43.4}} & {\color{myblue}\textbf{14.6}} &  17.7 & 43.6 & 90.9 & -  & 60.6\\
    \midrule
    Uni3D-MoE (ours) & {\color{myred}\textbf{30.8}}  & {\color{myred}\textbf{43.7}} & {\color{myred}\textbf{17.5}} & {\color{myblue}\textbf{19.0}} & {\color{myblue}\textbf{47.1}} & {\color{myred}\textbf{97.6}} & {\color{myred}\textbf{57.2}} & {\color{myred}\textbf{85.2}} \\
      \bottomrule
    \end{tabular}
    }
    \label{tab:results-scanqa}
\end{table}

\subsection{Analysis of MoE}
\begin{figure}
    \centering
    \includegraphics[width=1.0\linewidth]{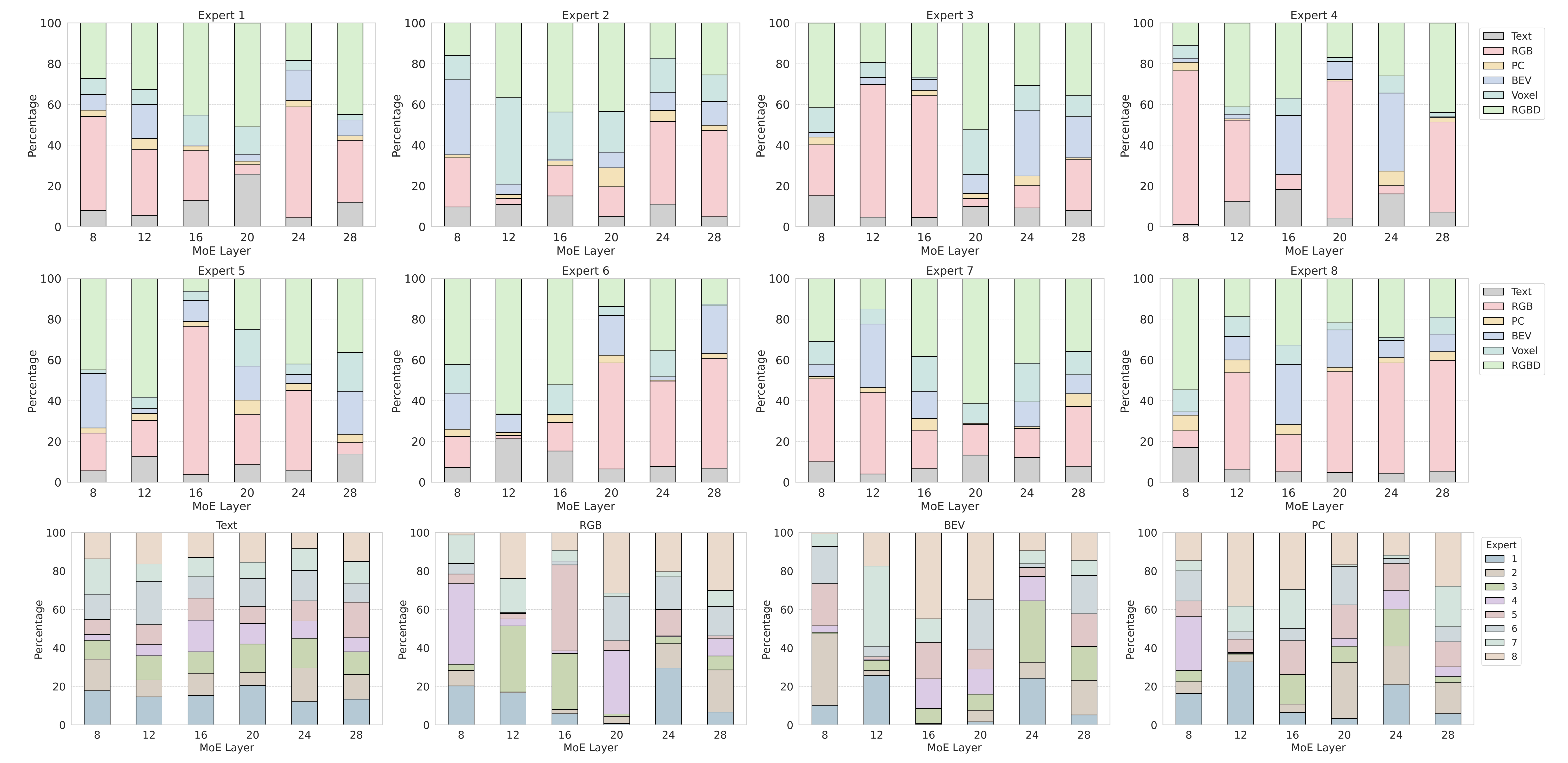}
    \caption{Token-to-expert routing across MoE layers. The first two rows show the modality token distribution across different experts at various MoE layers. The higher proportion of RGB, RGBD, and BEV tokens is attributed to their larger token counts, while expert-wise distributions reveal each expert's modality preferences. The third row presents the expert assignment distribution for each modality, indicating how each modality tends to select different experts throughout the MoE layers.}
    \label{fig:route}
\end{figure}
\textbf{Modality Aware Expert Specialization.}
Figure~\ref{fig:route} illustrates the modality token routing behavior across MoE layers from two perspectives: expert-centric and modality-centric.
First, the top two rows of Fig.~\ref{fig:route} visualize the distribution of modality tokens across 8 experts at each MoE layer.
The high proportions of RGB, RGBD, and BEV tokens are attributed to their larger token counts in the input. The expert-wise distributions still capture each expert’s modality specialization and selection preferences.
For instance, expert $\mathcal{E}_2$ shows a tendency to process voxel and point cloud tokens, which may indicate a specialization in geometric and structural information.
Expert $\mathcal{E}_4$ tends to focus more on RGB and BEV inputs, suggesting a possible strength in handling appearance and spatial-view representations.
Additionally, expert $\mathcal{E}_6$ at layer 12 and expert $\mathcal{E}_6$ at layer 20 display a noticeable preference for RGBD, potentially reflecting an ability to integrate color and depth cues.
Second, the third row of Fig.~\ref{fig:route} presents a modality-centric view of expert routing, reflecting which experts usually process tokens of this modality.
The results also suggests that multi-view RGB tokens are more often processed by expert $\mathcal{E}_4$, while point cloud tokens tend to be handled by experts $\mathcal{E}_2$, $\mathcal{E}_7$, and $\mathcal{E}_8$.
Text tokens appear more evenly distributed across all experts, which may indicate that each expert possesses a basic capacity for processing language information.
Overall, the results highlight the effectiveness of our MoE design in promoting modality-aware expert specialization.

\begin{wrapfigure}{r}{0.45\linewidth}
    \centering
    \vspace{-1em}
    \includegraphics[width=\linewidth]{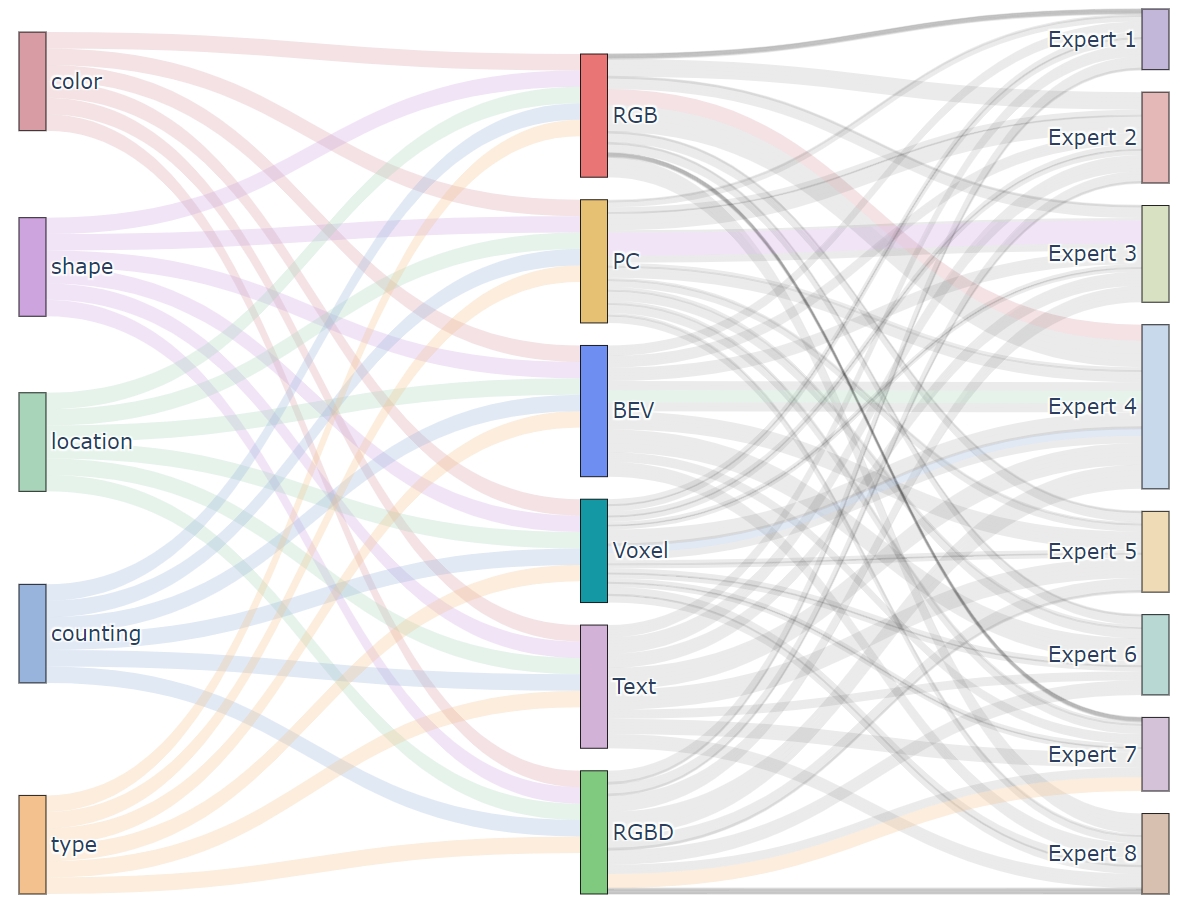}
    \caption{Modality-expert routing preferences across different question types. Line thickness indicates normalized token routing proportions. Preferred modality-expert routes for each query type are highlighted in color; others are shown in gray.}
    \label{fig:qtype}
    \vspace{-1em}
\end{wrapfigure}
\textbf{Question-type Aware Expert Specialization.}
To explore expert-specific modality preferences across question types, we select 5 representative categories (color, shape, location, counting, and type), each containing 500 samples from downstream validation/test sets.
To mitigate biases caused by differing token counts among modalities, token routing frequencies were normalized within each modality. 
Fig.~\ref{fig:qtype} provides insights into modality-expert routing preferences across these question categories.
Line thickness indicates normalized token routing proportions. Preferred modality-expert routes for each question type are highlighted in color; others are shown in gray.
For instance, color-related questions show a preference for the RGB modality tokens routed predominantly to expert $\mathcal{E}_4$, and shape-related questions favor the point cloud modality routed primarily to expert $\mathcal{E}_3$.
Additionally, expert $\mathcal{E}_4$ consistently exhibits relatively high activations across multiple question categories, potentially indicating its broader applicability within the MoE structure.
These observations demonstrate that our model’s routing mechanism exhibits adaptive modality preferences tailored to the specific question.

\begin{figure}[htbp]
    \centering
    \begin{minipage}[t]{0.52\textwidth}
        \centering
        \includegraphics[width=\linewidth]{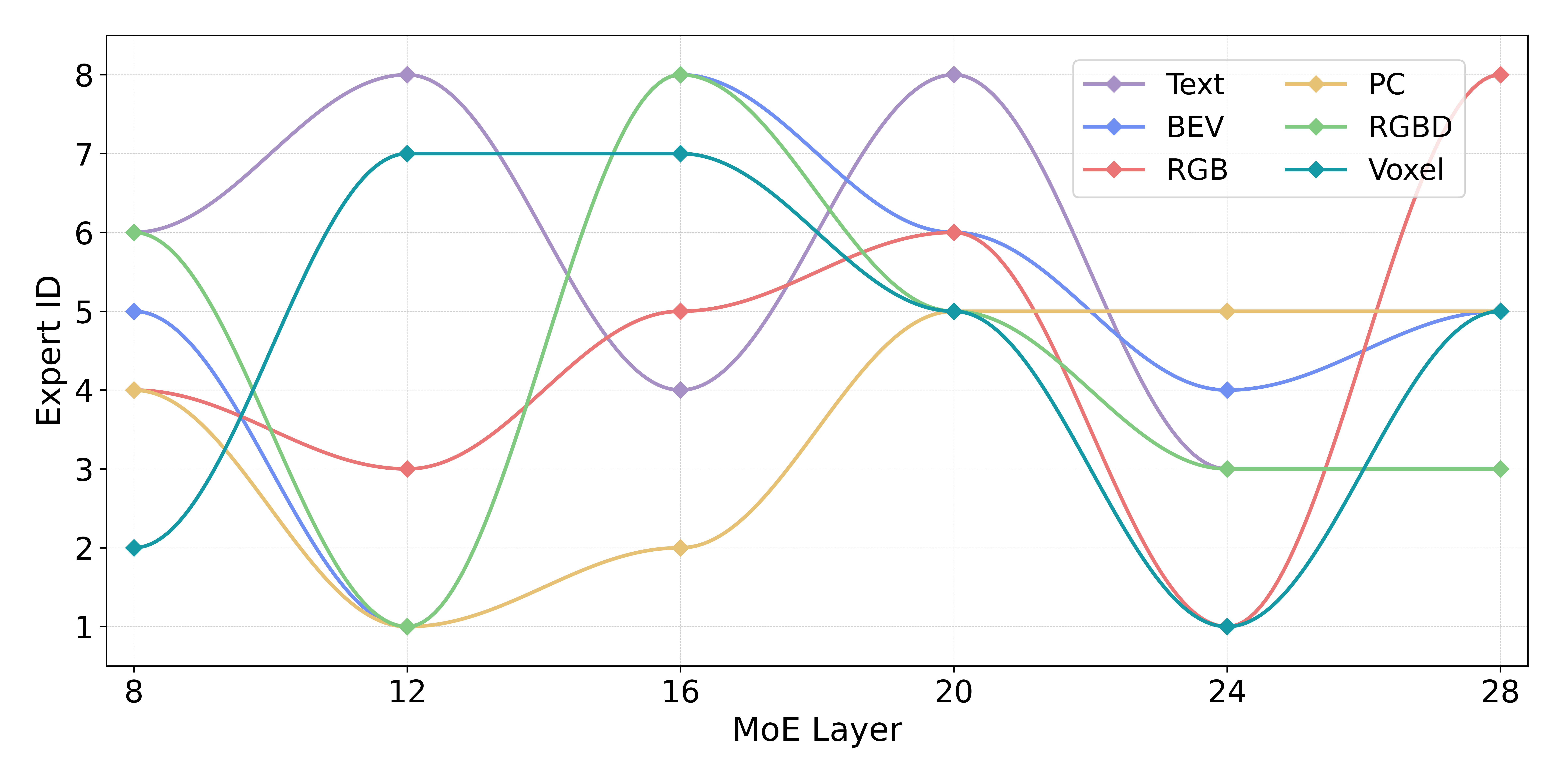}
        \vspace{-1em}
        \caption{Top-1 activated routing pathways for different modalities, highlighting dynamic and specialized expert activation.}
        \label{fig:route-path}
    \end{minipage}%
    \hfill
    \begin{minipage}[t]{0.43\textwidth}
        \centering
        \includegraphics[width=\linewidth]{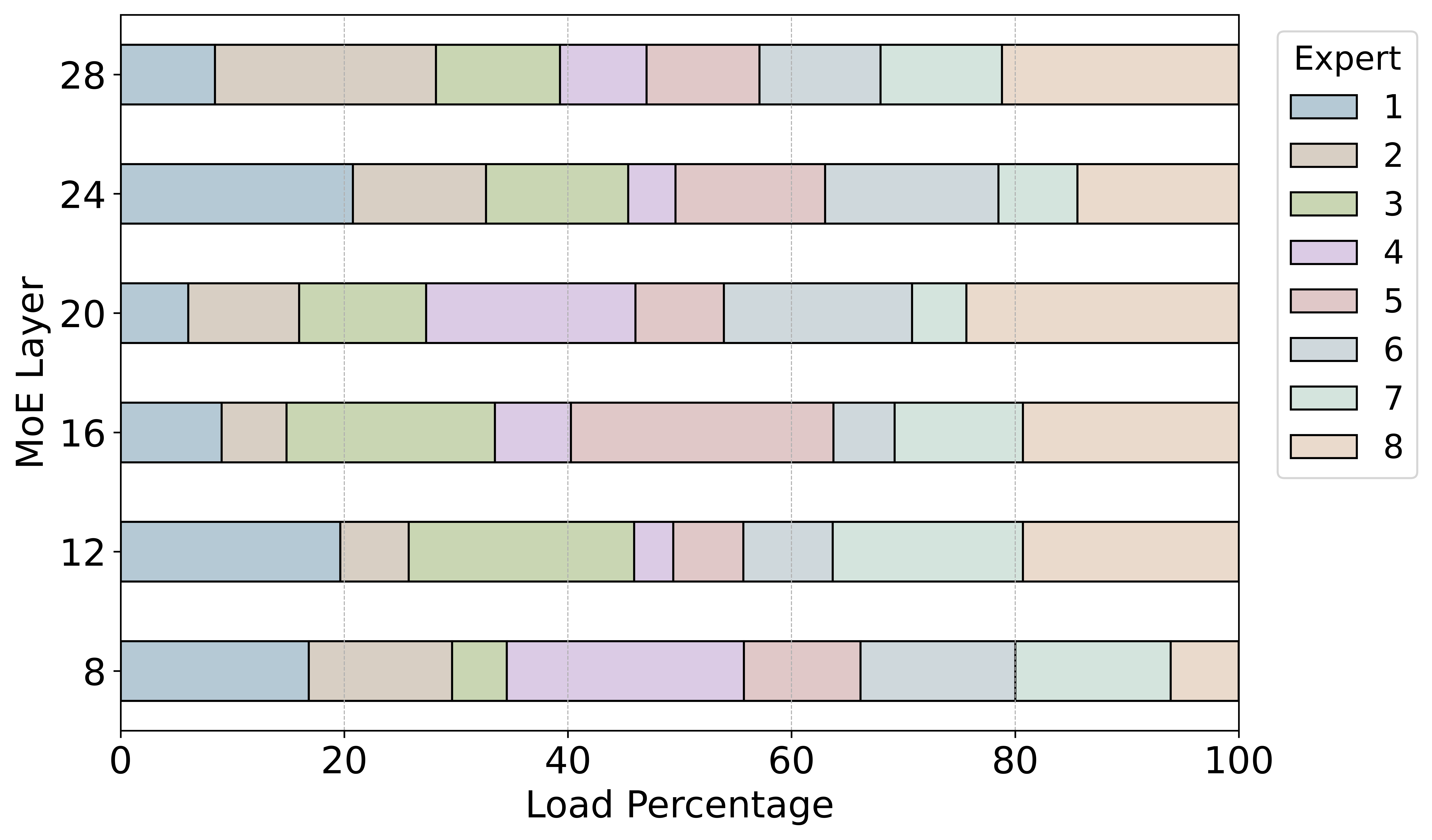}
        \vspace{-1em}
        \caption{The proportion of tokens assigned to each expert indicates the overall load balance and routing diversity.}
        \label{fig:route-load}
    \end{minipage}
    \vspace{-2em}
\end{figure}

\textbf{Activated Routing Pathways.}
We further track all tokens on downstream tasks and apply PCA~\citep{mackiewicz1993principal} to identify the top-10 most representative routing pathways. Fig.~\ref{fig:route-path} visualizes the top-1 activated expert routing trajectories across MoE layers for each modality. 
Please refer to the Appendix for more visualization results.
We observe that, at layer 20, previously unseen RGB and BEV tokens tend to be routed to expert $\mathcal{E}_6$, whereas PC, voxel, and depth (RGBD) tokens prefer expert $\mathcal{E}_5$. 
This suggests that Uni3D-MoE may implicitly distinguish visual semantic information from 3D geometric information during high-level representation learning, assigning them accordingly to suitable experts.
Additionally, PC and voxel tokens consistently share the same expert across layers 12, 20, and 28, which, to some extent, indicates the model's stable preference for spatial structural features.
Overall, these results highlight Uni3D-MoE's modality-aware dynamic routing, enhancing multimodal fusion for scene understanding.

\textbf{Expert Load Balance.}
Figure~\ref{fig:route-load} shows the proportion of tokens assigned to each expert across MoE layers.
Benefiting from the sparsity-aware expert balancing loss $l_{moe}$, the overall token distribution remains relatively balanced, which helps improve routing diversity and prevents expert under-utilization.


\subsection{Ablation Study}
In this section, we conduct ablation studies on Uni3D-MoE.
First, we evaluate modality contributions to various question categories.
Then, we examine the effectiveness of the MoE module.

\textbf{Ablation on Modality Contribution.}
As shown in Table~\ref{tab:ab-modality}, each modality contributes distinctly to different question types. 
Removing multi-view RGB modality (w/o $\mF_{rgb}$) primarily impacts performance on the ``color'' questions, indicating its crucial role in capturing visual color details. 
Excluding the BEV modality (w/o $\mF_{bev}$) notably reduces performance on ``Location'' and ``type'' tasks, highlighting its importance in spatial understanding. 
Omitting RGB-D modality (w/o $\mF_{rgbd}$) mainly decreases performance in the ``nature'' category, indicating its essential role in capturing detailed characteristics such as object type and shape. 
Removing point cloud data (w/o $\mF_{pc}$) considerably affects performance on both ``counting'' and ``nature'' categories, demonstrating its strength in capturing object count and structural features. 
The ablation of voxel modality (w/o $\mF_{voxel}$) leads to a notable performance drop in ``location'' and ``counting'' tasks, underscoring its effectiveness in detailed spatial and quantity understanding.
Overall, these results clearly illustrate the complementary and task-specific roles of each modality in Uni3D-MoE, highlighting how their specialized cues collectively facilitate accurate semantic understanding and effective question-specific reasoning.

\textbf{Ablation on MoE.}
First, we evaluate the effectiveness of the MoE module across different question categories.
Specifically, we select and construct five types of questions (location, type, nature, counting, and color) from downstream tasks. 
Table~\ref{tab:ab-moe-type} demonstrates that incorporating the MoE module consistently improves performance across all these categories, with notable CIDEr gains for ``type'' and ``color''.
Then, we investigate the effect of varying the number of experts within MoE layers.
Considering GPU memory constraints, we set the number of experts per MoE layer to 4, 6, and 8, with each token routed to the top-2 experts. 
As shown in Table~\ref{tab:ab-moe-number}, the model with 8 experts per MoE layer outperforms the baseline without MoE by 9.2 on CIDEr.
Furthermore, the results show a trend of improved performance with an increasing number of experts, though at the cost of increased training time, suggesting a trade-off between accuracy and computational cost. Additional ablations on the MoE module, including the choice of MoE-equipped layers, are provided in the Appendix.

\begin{table}[t!]
    \centering
    \small
    \caption{Ablation results of input modalities across five question categories, including location, type, nature, counting, and color, where ``nature'' mainly covers object shape and type.  ``w/'' denotes experiments using only the specified modality; ``w/o'' denotes experiments excluding that modality. ``w/o 2D info'': only point cloud and voxel; ``w/o 3D info'': only RGB and BEV.}
    \resizebox{\textwidth}{!}{
    \begin{tabular}{lcccccccccc}
        \toprule
        \multirow{2}{*}{\textbf{Method}} &  \multicolumn{2}{c}{\textbf{Location}} & \multicolumn{2}{c}{\textbf{Type}} & \multicolumn{2}{c}{\textbf{Nature}} & \multicolumn{2}{c}{\textbf{Counting}} & \multicolumn{2}{c}{\textbf{Color}}\\
        \cmidrule(lr){2-3} \cmidrule(lr){4-5} \cmidrule(lr){6-7} \cmidrule(lr){8-9} \cmidrule(lr){10-11} 
        & F1  & CIDEr & F1 & CIDEr & F1 & CIDEr & F1  & CIDEr & F1  & CIDEr \\
        \midrule
      \rowcolor{green!5}
         w/ $\mF_{rgb}$ &  25.48  & 46.42 & 22.89 & 40.37 & 41.64 & 76.38 &  45.51 & 68.03 & 37.43 &  62.47  \\
      \rowcolor{green!5}
        w/ $\mF_{rgbd}$ &  20.79 & 31.50 & 17.27 & 31.54 & 32.34 & 60.65 & 37.95 & 59.66 & 36.96 & 57.43\\
        \rowcolor{green!5}
          w/ $\mF_{bev}$ & 19.92 & 28.76 & 18.35 & 33.56 & 31.97 & 56.33 & 38.94 & 62.89 & 33.65 &  56.92 \\
                \rowcolor{green!5}
         w/ $\mF_{pc}$ &  18.71 & 26.26 & 16.95 & 28.34 & 33.62 & 59.31 & 38.54 & 62.83 & 36.73 & 62.31 \\
               \rowcolor{green!5}
        w/ $\mF_{voxel}$ &  16.00 & 25.39 &  18.20 & 31.57 & 37.02 & 69.35 & 42.12  & 70.09 &  31.76 & 56.85\\
        \midrule
        \rowcolor{yellow!5}
         w/o $\mF_{rgb}$ & 32.42 &  59.89 & 31.01 & 55.92 & 52.25 & 102.53 & 53.08 & 89.53 & 49.78 & 86.52  \\ 
         \rowcolor{yellow!5}
          w/o $\mF_{rgbd}$ &  34.67 & 66.34 &  31.93 & 55.41 & 51.25 & 97.31 & 51.39 & 77.83 & 50.32 & 87.86   \\
          \rowcolor{yellow!5}
          w/o $\mF_{bev}$ &  32.43 & 60.17 & 30.94 & 54.85 & 51.58 & 98.58 & 52.65 & 83.41 & 49.16 & 85.75  \\
          \rowcolor{yellow!5}
         w/o $\mF_{pc}$ &  35.06 & 69.19 &  31.73 & 55.88 & 50.92 & 96.86 & 49.68 & 75.20 & 51.59 & 89.30 \\
         \rowcolor{yellow!5}
        w/o $\mF_{voxel}$ & 33.16 & 59.46 & 32.54 & 59.21  & 51.68 & 99.71 & 38.58 & 59.69 &  51.20 & 88.68   \\
         \midrule
         \rowcolor{blue!5}
         w/o 2D info & 34.33 & 61.63 & 30.47 & 55.31 & 48.61 & 91.40 & 49.82 & 77.64 & 48.64 & 83.40 \\
         \rowcolor{blue!5}
         w/o 3D info & 34.33 & 65.57 & 31.22 & 55.31 & 48.25 & 89.41 & 50.33 & 77.95 & 51.54 & 90.80 \\ 

         \midrule
         \rowcolor{red!5}
         w/ $\mathcal{F}_{uni}$ & 37.57 & 78.59 & 42.68 & 80.32 & 52.89 & 106.40 & 50.59 &  79.52 & 62.52 & 111.08   \\         
         \bottomrule
    \end{tabular}
    }
    \label{tab:ab-modality}
\end{table}

\begin{table}[t!]
    \centering
    \small
    \caption{Ablation results of MoE module across five question categories, including location, type, nature, counting, and color, where ``nature'' mainly covers object shape and type.}
    \resizebox{\textwidth}{!}{
    \begin{tabular}{lllllllllll}
        \toprule
        \multirow{2}{*}{\textbf{Method}} &  \multicolumn{2}{c}{\textbf{Location}} & \multicolumn{2}{c}{\textbf{Type}} & \multicolumn{2}{c}{\textbf{Nature}} & \multicolumn{2}{c}{\textbf{Counting}} & \multicolumn{2}{c}{\textbf{Color}}\\

        \cmidrule(lr){2-3} \cmidrule(lr){4-5} \cmidrule(lr){6-7} \cmidrule(lr){8-9} \cmidrule(lr){10-11} 
        & F1  & CIDEr & F1 & CIDEr & F1 & CIDEr & F1  & CIDEr & F1  & CIDEr \\
        \midrule
        w/o MoE & 37.5 & 78.5 & 42.6 & 80.3 & 52.8 & 106.4 & 50.5 &  79.5 & 62.5 & 111.0   \\ 
        \rowcolor{red!5}
        w/ MoE  & 39.5$_{\color{red}\uparrow2.0}$ & 85.1$_{\color{red}\uparrow6.6}$  & 47.3$_{\color{red}\uparrow4.7}$ & 93.7$_{\color{red}\uparrow13.4}$ & 53.4$_{\color{red}\uparrow0.6}$ & 107.9$_{\color{red}\uparrow1.5}$ & 53.9$_{\color{red}\uparrow3.4}$ & 86.3$_{\color{red}\uparrow6.8}$ & 67.6$_{\color{red}\uparrow5.1}$ & 121.2$_{\color{red}\uparrow10.2}$ \\
        
         \bottomrule
    \end{tabular}
    }
    \label{tab:ab-moe-type}
\end{table}

\begin{table}[t!]
    \centering
    \small
    \caption{Ablation results on the number of experts ($E$) in the MoE module on ScanQA\citep{azuma2022scanqa}. ``w/o MoE'' indicates the baseline without MoE. ``Time'' indicates second-stage MoE training duration.}
    \resizebox{\textwidth}{!}{
    \begin{tabular}{llllllllllll}
        \toprule
        \textbf{Method} & \textbf{E} & \textbf{EM@1} & \textbf{EM-R@1} & \textbf{F1} &  \textbf{BLEU-1} & \textbf{BLEU-4} & \textbf{METEOR} & \textbf{ROUGE} & \textbf{CIDEr} & \textbf{Time}\\
        \midrule
         w/o MoE & - & 27.3 & 45.1 & 45.5 & 41.9 & 13.9 & 17.1 & 43.8 &  88.4 & - \\ 
         w/ MoE & 4 & 29.5 & 47.1 & 47.3 & 43.2 & 15.8 & 18.2 & 45.7 & 93.1 & $\sim12h$ \\
         w/ MoE & 6 & 29.9 & 47.8 & 48.4 & 43.2 &  16.4 & 19.0 & 45.9 & 95.5 & $\sim14h$\\
         \rowcolor{red!5}
         w/  MoE & 8 & 30.8$_{\color{red}\uparrow3.5}$ & 49.0$_{\color{red}\uparrow3.9}$ & 48.8$_{\color{red}\uparrow3.3}$ & 43.7$_{\color{red}\uparrow1.8}$ & 17.5$_{\color{red}\uparrow3.6}$ & 19.0$_{\color{red}\uparrow1.9}$ & 47.1$_{\color{red}\uparrow3.3}$ & 97.6$_{\color{red}\uparrow9.2}$ & $\sim17h$ \\
         \bottomrule
    \end{tabular}
    }
    \label{tab:ab-moe-number}
\end{table}

\subsection{Limitations}\label{sec:exper-limit}
Despite promising results, Uni3D-MoE exhibits limitations due to token budget constraints and dataset quality. Token limits necessitate modality tokens reduction strategies: multi-view images selected by MVCS may omit critical viewpoints, causing incomplete spatial context; similarly, FPS downsampling reduces point cloud density, compromising fine-grained details. Additionally, performance is affected by blurry images and annotation inaccuracies, introducing noise that impacts precise spatial understanding and object localization tasks.

\section{Conclusion}
In this paper, we propose Uni3D-MoE, a MoE-based 3D MLLM for comprehensive and adaptive scene understanding.
Uni3D-MoE integrates multi-view images, depth, BEV, point clouds, and voxels through modality-specific encoders and a sparse MoE mechanism.
By augmenting the LLM with learnable sparse MoE layers, our model adaptively activates specialized experts tailored to each token, enabling dynamic modality fusion aligned with prompt-specific needs.
Experimental results show that our method achieves competitive performance on multiple scene understanding tasks.

\small
\bibliographystyle{unsrt}
\bibliography{reference}
\medskip

\clearpage
\appendix

\section{Summary}\label{appendix:summary}
The appendix is organized as follows:

\textbf{Appendix~\ref{appendix:uni3d-results}:More Results of Uni3D-MoE.} \\
Appendix~\ref{appendix:visual-qualitative}: Visualization Results. \\
Appendix~\ref{appendix:visual-failure}: Failure Cases.  \\
Appendix~\ref{appendix:quant-results}: Detailed quantitative comparison between Uni3D-MoE and other baseline models on various benchmarks, including ScanQA~\citep{azuma2022scanqa}, SQA3D~\citep{ma2022sqa3d}, ScanRefer~\citep{chen2020scanrefer}, Multi3DRefer~\citep{zhang2023multi3drefer} and Scan2Cap~\citep{chen2021scan2cap}.  \\
\textbf{Appendix~\ref{appendix:moe-results}: Additional Results for the MoE Module
.} \\
Appendix~\ref{appendix:moe-vidual}: Visualization results of MoE, including top-10 activated routing pathways and expert assignment distribution for each modality. \\
Appendix~\ref{appendix:moe-abl}: Ablation results of MoE, including experiments on MoE layer placement and evaluations across multiple benchmarks. \\
\textbf{Appendix~\ref{appendix:limitations+bi}: Limitations and Broader Impacts.}  \\
\textbf{Appendix~\ref{appendix:data}: Data Details}, including dialogue data format and prompt template. \\
\textbf{Appendix~\ref{appendix:model-details}: Model Details,} including modality-specific encoders, adapters, and the sparse MoE-based LLM. \\

\section{More Results of Uni3D-MoE}\label{appendix:uni3d-results}
\subsection{Visualization Results}\label{appendix:visual-qualitative}
\begin{figure}[h!]
    \centering
    \includegraphics[width=1.0\linewidth]{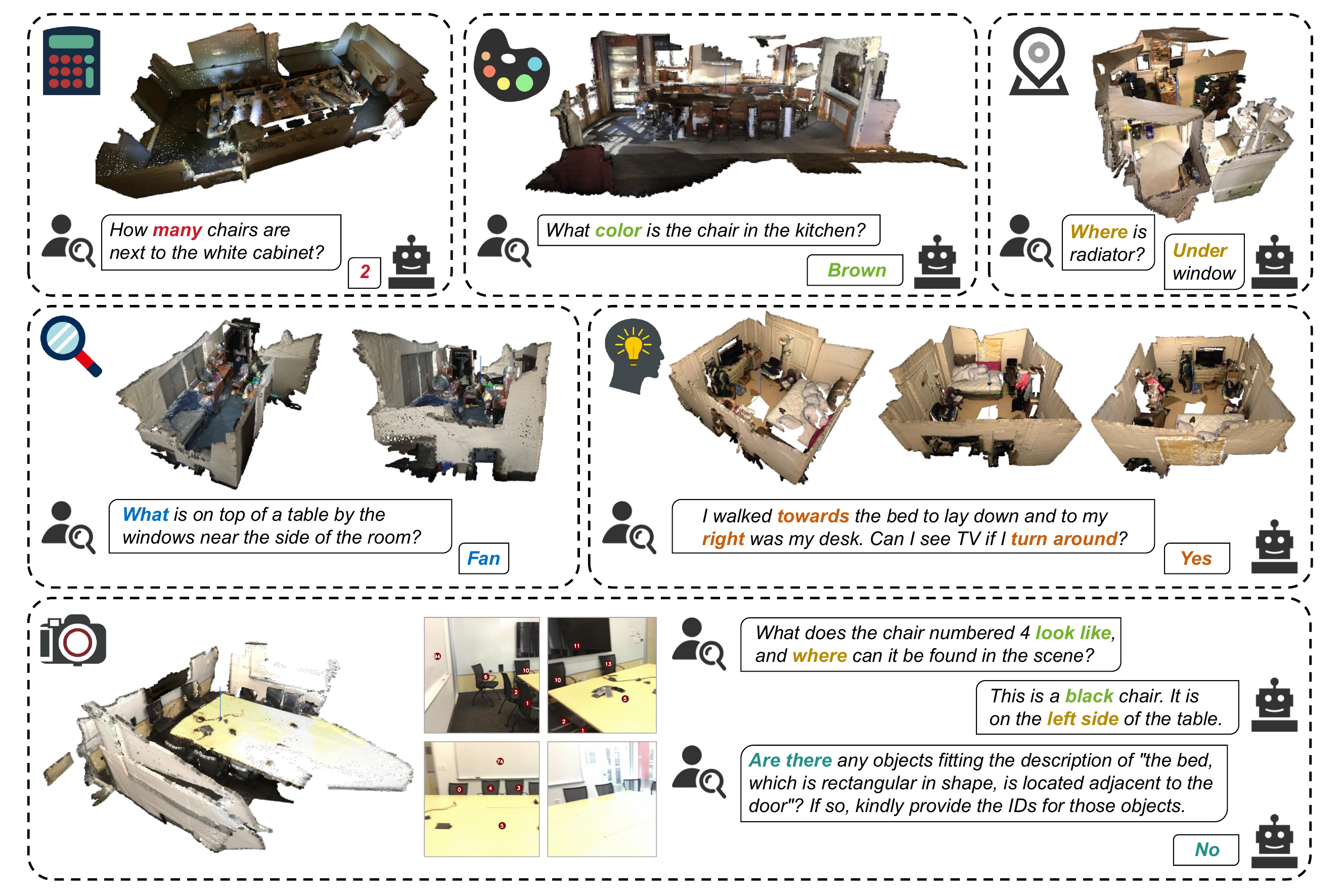}
    \caption{Visualization of Uni3D-MoE performing diverse 3D scene understanding tasks. Examples highlight the model's adaptive multimodal reasoning capabilities across different query types including counting, color recognition, object localization, spatial reasoning, and semantic identification.}
    \label{fig:chat}
\end{figure}
As illustrated in Fig.~\ref{fig:chat}, the proposed Uni3D-MoE model adeptly handles various categories of 3D scene understanding tasks, effectively demonstrating its versatility across distinct question types. For instance, it accurately identifies numeric details in counting tasks (e.g., "How many chairs are next to the white cabinet?"), utilizes color recognition to specify attributes (e.g., the "brown" chair in the kitchen), and spatially localizes objects by contextual information (e.g., finding a radiator "under the window"). Moreover, the model is capable of interpreting spatial orientation and viewpoint-dependent questions, successfully answering queries related to turning around to view specific objects. Conversely, it can clearly recognize when queried objects or conditions are absent in the scene, indicating robust negative reasoning capability. These diverse examples underscore Uni3D-MoE's effectiveness in dynamically leveraging multimodal data representations, showcasing its capability for nuanced and contextually adaptive responses.

\subsection{Failure Cases}\label{appendix:visual-failure}
\begin{figure}[h!]
    \centering
    \includegraphics[width=\linewidth]{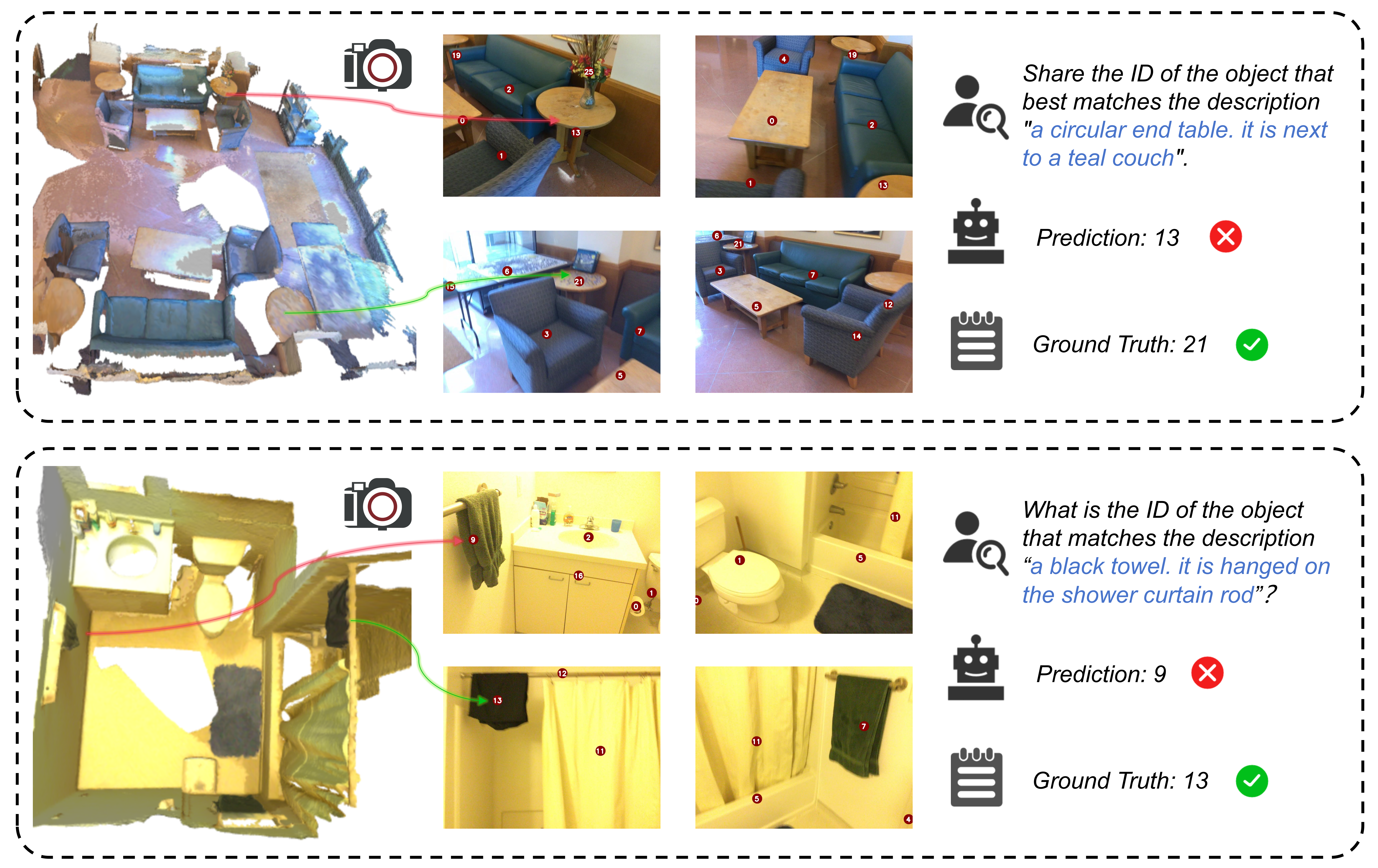}
    \caption{Failure cases of Uni3D-MoE.}
    \label{fig:appendix-visual-failure}
\end{figure}

Fig.~\ref{fig:appendix-visual-failure} illustrates the failure cases of our method.
In the first example, given the query ``a circular end table, it is next to a teal couch'' the model predicts object 13, while the ground truth is object 21.
Notably, object 13 also accurately satisfies the description, as it is similarly positioned next to a teal couch and matches the described shape. 
This indicates that the error arises primarily from inherent annotation ambiguity, rather than from a fundamental shortcoming of the model’s visual grounding capability. 
In the second example, for the query ``a black towel, it is hung on the shower curtain rod'' the model incorrectly selects object 9 instead of the correct object 13. 
This failure may be attributed to inconsistent lighting conditions between this scene and others, resulting in color deviations in multi-view RGB images, thus impairing the model's ability to accurately interpret visual cues and distinguish subtle color differences.
Additionally, the incorrect prediction might stem from the higher occurrence frequency of object 9 across multiple frames, potentially biasing the model’s attention toward it over the less prominently featured yet correct object 13. 
These cases highlight the importance of addressing both annotation ambiguity and robustness to visual variations in future model improvements.

\subsection{Quantitative Comparison Results}\label{appendix:quant-results}

\textbf{Compared Baselines.}\label{appendix:baseline}
We comprehensively evaluate Uni3D-MoE against three categories of state-of-the-art approaches across several 3D benchmarks: 
\begin{itemize}
    \item \textit{Task-specific models} specifically optimized for individual 3D tasks, including ScanQA~\citep{azuma2022scanqa}, 3D-VLP~\citep{jin2023context}, 3D-VisTA~\citep{zhu20233d}, Scan2Cap~\citep{chen2021scan2cap}, 3DJCG~\citep{cai20223djcg}, Vote2Cap-DETR~\citep{chen2023end}, X-Trans2Cap~\citep{yuan2022x}, ScanRefer~\citep{chen2020scanrefer}, MVT~\citep{chen2021mvt}, 3DVG-Trans~\citep{zhao20213dvg}, ViL3DRel~\citep{chen2022language}, and M3DRef-CLIP~\citep{zhang2023multi3drefer}.
    \item 
    \textit{2D LLMs} adapted from general image-based vision-language models, such as InternVL2-8B~\citep{lu2025internvl}, Qwen2-VL-7B~\citep{wang2024qwen2}, and LLaVA-Video~\citep{zhang2024video}.
    \item 
    \textit{3D LLMs} that integrate multimodal 3D information into pretrained language models, including PQ3D~\citep{zhu2024unifying}, LAMM~\citep{yin2023lamm}, Chat-3D~\citep{wang2023chat}, Chat-3D V2~\citep{DBLP:journals/corr/abs-2312-08168}, 3D-LLM~\citep{hong20233d}, LL3DA~\citep{chen2024ll3da}, LEO~\citep{huang2024embodied}, Scene-LLM~\citep{fu2024scene}, Chat-Scene~\citep{huang2024chat}, LLaVA-3D~\citep{zhu2024llava}, GPT4Scene~\citep{qi2025gpt4scene}, Ground 3D-LLM~\citep{chen2024grounded}. 
\end{itemize}

\textbf{Metric Details.}\label{appendix:metric}
Following previous work~\citep{zhu2024llava, huang2024chat, wang2025ross3d}, we comprehensively evaluate our method using standard metrics across multiple tasks in 3D scene understanding. Specifically:
\begin{itemize}
    \item 
    For the Scan2Cap~\citep{chen2021scan2cap} task, we assess the quality of generated scene descriptions using widely adopted captioning metrics, including BLEU-4, METEOR, ROUGE, and CIDEr, computed specifically at Intersection-over-Union (IoU) thresholds of 0.25 and 0.5 between predicted and ground-truth bounding boxes.
    \item 
    For the ScanQA~\citep{azuma2022scanqa} question-answering task, besides captioning metrics, we utilize metrics tailored for answer accuracy and completeness: Exact Match accuracy (EM@1) measures strict correctness of top-1 answers, Relaxed Exact Match (EM-R@1) allows minor acceptable variations, and F1 scores evaluate token-level overlaps.
    \item 
    For referring expression grounding tasks, ScanRefer~\citep{chen2020scanrefer} and Multi3DRefer~\citep{zhang2023multi3drefer}, we evaluate localization accuracy of predicted bounding boxes against ground-truth annotations. Specifically, we report accuracy (Acc@0.25, Acc@0.5) at IoU thresholds of 0.25 and 0.5 for ScanRefer, and F1 scores (F1@0.25, F1@0.5) at the same IoU thresholds for Multi3DRefer.
\end{itemize}

In summary, the metrics used can be grouped into three categories: text similarity metrics (BLEU, METEOR, ROUGE, CIDEr) for assessing the quality and fluency of generated descriptions; accuracy metrics (EM@1, EM-R@1, F1) for evaluating exactness and completeness in question-answering tasks; and spatial localization metrics (Acc@IoU, F1@IoU, captioning metrics at IoU thresholds) to quantify the accuracy of bounding-box predictions in scene grounding tasks.

\textbf{3D Visual Question Answering.}
Table~\ref{tab:appendix-qt-sqa3d} provides a comprehensive evaluation of various models on the SQA3D benchmark across different 3D question-answering tasks. The tasks are categorized by question types, including ``What'', ``Is'', ``How'', ``Can'', ``Which'', and ``Others'', alongside aggregated metrics of Exact Match accuracy (EM@1) and Relaxed Exact Match accuracy (EM-R@1). Our method, Uni3D-MoE, demonstrates superior performance compared to existing state-of-the-art methods across multiple question types. Specifically, Uni3D-MoE achieves the best results on the "What" (53.1\%), "How" (55.8\%), "Which" (55.3\%), and "Others" (60.2\%) question categories, while obtaining second-best performance in ``Is'' (69.9\%). When examining overall accuracy metrics, our model achieves an EM@1 of 57.2\% and EM-R@1 of 59.8\%, outperforming most baseline methods significantly. These results suggest that our integration of multimodal Mixture-of-Experts (MoE) architecture effectively enhances the model’s capacity to process and interpret complex 3D scene queries, particularly in handling open-ended and detailed inquiries.

\begin{table}[h!]
    \centering
    \small
    \caption{Evaluation results of 3D question answering across different question types on the test set of SQA3D~\citep{ma2022sqa3d}. $\star$ indicates that high-resolution settings are not used. We highlight the best performance in {\color{myred}\textbf{red}} and the second-best in {\color{myblue}\textbf{blue}}.}
    \resizebox{\textwidth}{!}{
    \begin{tabular}{lccccccccc}
        \toprule
            \multirow{2}{*}{\textbf{Method}} & \multicolumn{6}{c}{\textbf{Question Type}}  & \multicolumn{2}{c}{\textbf{Total}}\\
             \cmidrule(lr){2-7}  \cmidrule(lr){8-9}
           &  What & Is & How & Can & Which & Others & EM@1 & EM-R@1 \\
        \midrule
\textbf{\textit{Task-specific}} \\
SQA3D~\citep{ma2022sqa3d} & 31.6  & 63.8  & 46.0  &69.5 &43.9  & 45.3 &46.6 & - \\
3D-VisTA~\citep{zhu20233d} & 34.8 & 63.3  & 45.4 & 69.8  & 47.2  & 48.1  &48.5 & - \\
ClipBERT~\citep{lei2021less} &  30.2 & 60.1 & 38.7 & 63.3 & 42.5 & 42.7 & 43.3 & – \\
\textbf{\textit{2D LLMs}} \\
InternVL2-8B~\citep{lu2025internvl} &  30.5  & 53.8 & 5.5 & 47.3  & 25.8  & 36.3 & 33.0 & 45.3  \\
Qwen2-VL-7B~\citep{wang2024qwen2} & 29.0 & 59.2 & 33.4 & 50.5 & 44.2 & 43.2 & 40.7 & 46.7 \\
LLaVA-Video-7B~\citep{zhang2024video} & 42.7 & 56.3 & 47.5 & 55.3 & 50.1 & 47.2 & 48.5 & – \\
\textbf{\textit{3D LLMs}}\\
LEO~\citep{huang2024embodied} & – & – & – & – & – & – & 50.0 & 52.4 \\
Scene-LLM~\citep{fu2024scene} & 40.9 & 69.1 & 45.0 & {\color{myred}\textbf{70.8}} & 47.2 & 52.3 & 54.2 & – \\
ChatScene~\citep{huang2024chat}  & 45.4 & 67.0 & 52.0 & 69.5 & 49.9 & 55.0 & 54.6 & 57.5 \\
LLaVA-3D~\citep{zhu2024llava} & – & – & – & –  &– &– & 55.6 & – \\
GPT4Scene$^{\star}$ ~\citep{qi2025gpt4scene} & {\color{myblue}\textbf{50.7}} & {\color{myred}\textbf{70.9}} & 48.0 & {\color{myblue}\textbf{70.5}} & {\color{myblue}\textbf{52.9}} & {\color{myblue}\textbf{59.3}} & - & {\color{myred}\textbf{60.7}}\\ 
\midrule
Ours & {\color{myred}\textbf{53.1}} & {\color{myblue}\textbf{69.9}} & {\color{myred}\textbf{55.8}} & 69.5 & {\color{myred}\textbf{55.3}} & {\color{myred}\textbf{60.2}}  & 57.2 & {\color{myblue}\textbf{59.8}}\\

 \bottomrule
    \end{tabular}
    }
    \label{tab:appendix-qt-sqa3d}
\end{table}

\textbf{3D Visual Grounding.}
Table~\ref{tab:appendix-scanrefer+multi3drefer} summarizes evaluation results for 3D visual grounding tasks on ScanRefer~\citep{chen2020scanrefer} and Multi3DRefer~\citep{zhang2023multi3drefer}, comparing task-specific models and general 3D Large Language Models (3D LLMs). Our method achieves state-of-the-art results, outperforming existing approaches on both benchmarks. Specifically, our model attains the highest accuracy of 62.7\% and 57.4\% at IoU thresholds of 0.25 and 0.5 respectively on ScanRefer~\citep{chen2020scanrefer}, and F1-scores of 65.1\% and 60.5\% at IoU thresholds of 0.25 and 0.5 on Multi3DRefer~\citep{zhang2023multi3drefer}, demonstrating improvements over the previous best-performing method, GPT4Scene-HDM~\citep{qi2025gpt4scene}. These results validate the efficacy of incorporating a Mixture-of-Experts (MoE) architecture into multimodal LLMs, highlighting substantial gains in multimodal grounding capability.

\begin{wraptable}{r}{0.6\textwidth}
    \centering
    \caption{Evaluation results of 3D visual grounding on ScanRefer~\citep{chen2020scanrefer} and Multi3DRefer~\citep{zhang2023multi3drefer}. $\star$ indicates that high-resolution settings are not used. We highlight the best performance in {\color{myred}\textbf{red}} and the second-best in {\color{myblue}\textbf{blue}}.}
    \resizebox{0.6\textwidth}{!}{
    \begin{tabular}{ccccccccc}
    \toprule
     \multirow{2}{*}{\textbf{Method}} &  \multicolumn{2}{c}{\textbf{ScanRefer}} &  \multicolumn{2}{c}{\textbf{Multi3DRefer}}    \\
     \cmidrule(lr){2-3} \cmidrule(lr){4-5} 
       & Acc@0.25 & Acc@0.5 & F1@0.25 & F1@0.5  \\
     \midrule
     \textbf{\textit{Task-specific Models}} \\
      ScanRefer~\citep{chen2020scanrefer}& 37.3 & 24.3 & – & – \\
     MVT~\citep{chen2021mvt} & 40.8 & 33.3 & – & – \\
     3DVG-Trans~\citep{zhao20213dvg} & 47.6 & 34.7 & – & 25.5 \\
     ViL3DRel~\citep{chen2022language} & 47.9 & 37.7 & – & – \\
     3DJCG~\citep{cai20223djcg} & 49.6 & 37.3 & – & 26.6 \\
      M3DRef-CLIP~\citep{zhang2023multi3drefer} & 51.9 & 44.7 & 42.8 & 38.4 \\
      \midrule
      \textbf{\textit{3D LLMs}}\\
    3D-LLM~\citep{hong20233d} & 30.3 & – &  – & – \\
    Ground 3D-LLM~\citep{chen2024grounded} & 47.9 & 44.1 & 45.2 & 40.6 \\
    Chat-Scene~\citep{huang2024chat} & 55.5 & 50.2 & 57.1 & 52.4 \\
    LLaVA-3D~\citep{zhu2024llava} & 50.1 & 42.7  & – & – \\
    Ross3D~\citep{wang2025ross3d} & 61.1 & 54.4 & 59.6 & 54.3 \\
    GPT4Scene$^{\star}$~\citep{qi2025gpt4scene} & 40.5 & 36.7 & 45.4 & 42.1 \\
    GPT4Scene-HD~\citep{qi2025gpt4scene} & 50.9 &  46.4 & 53.7 & 50.0  \\
    GPT4Scene-HDM~\citep{qi2025gpt4scene} & {\color{myblue}\textbf{62.6}} & {\color{myblue}\textbf{57.0}} & {\color{myblue}\textbf{64.5}} & {\color{myblue}\textbf{59.8}}  \\
    
    \midrule
   Ours  & {\color{myred}\textbf{62.7}} & {\color{myred}\textbf{57.4}} &  {\color{myred}\textbf{65.1}} & {\color{myred}\textbf{60.5}} \\ %
      \bottomrule
    \end{tabular}
    }
    \label{tab:appendix-scanrefer+multi3drefer}
\end{wraptable}

Table~\ref{tab:appendix-scanrefer-detail} presents a comprehensive comparison between our method and other state-of-the-art approaches on the ScanRefer~\citep{chen2020scanrefer}.
Performance is assessed separately across ``Unique'', ``Multiple'', and combined ``Overall'' subsets. The ``Unique'' subset involves unambiguous samples, each with only a single instance per object category, whereas the ``Multiple'' subset includes ambiguous samples containing multiple instances from the same category. Metrics used are accuracy measured at IoU thresholds of 0.25 and 0.5.
The proposed method achieves superior performance, especially in handling ambiguous cases within the ``Multiple'' subset, obtaining promising accuracy scores at 56.7\% (Acc@0.25) and 51.5\% (Acc@0.5). It also demonstrates outstanding overall capabilities, achieving state-of-the-art results on the ``Overall'' subset with accuracies of 62.7\% and 57.4\%, closely surpassing the previously best-performing model GPT4Scene-HDM. In the ``Unique'' subset, our method achieves competitive results (89.6\% at Acc@0.25 and 83.5\% at Acc@0.5), second only slightly to GPT4Scene-HDM, reflecting strong capability in handling clear, well-defined visual grounding scenarios.
These results highlight substantial effectiveness of our method in visual grounding, notably its capability in resolving ambiguity inherent in challenging multi-instance scenes, thus underscoring the advantages brought by integrating Mixture-of-Experts architecture within multimodal large language models.

\begin{table}[h!]
    \centering
    \caption{Full Evaluation of 3D visual grounding on ScanRefer~\citep{chen2020scanrefer}. The ``Unique'' subset contains samples in which the described object corresponds to exactly one unique instance within a given object category, whereas the ``Multiple'' subset includes ambiguous cases with multiple instances belonging to the same object category. The ``Overall'' category aggregates performance across both unique and multiple-instance subsets. Accuracy is measured using IoU thresholds of 0.25 and 0.5 between predicted and ground-truth bounding boxes.  $\star$ indicates that high-resolution settings are not used. We highlight the best performance in {\color{myred}\textbf{red}} and the second-best in {\color{myblue}\textbf{blue}}.}
    \resizebox{\textwidth}{!}{
    \begin{tabular}{ccccccccccc}
    \toprule
     \multirow{2}{*}{\textbf{Method}} &  \multicolumn{2}{c}{\textbf{Unique}} &  \multicolumn{2}{c}{\textbf{Multiple}}  & \multicolumn{2}{c}{\textbf{Overall}}   \\
     \cmidrule(lr){2-3} \cmidrule(lr){4-5} \cmidrule(lr){6-7} 
       & Acc@0.25 & Acc@0.5 & Acc@0.25 & Acc@0.5 & Acc@0.25 & Acc@0.5  \\
     \midrule
\textbf{\textit{Task-specific Models}} \\
ScanRefer~\citep{chen2020scanrefer} & 76.3 & 53.5 & 32.7 & 21.1 & 41.2 & 27.4 \\
TGNN~\citep{huang2021text} & 68.6  & 56.8 & 29.8 & 23.2 & 37.4 & 29.7 \\
X-Trans2Cap~\citep{yuan2022x} & 73.2 & 50.8 & 37.6 & 25.2 & 44.5 & 30.1 \\
InstanceRefer~\citep{yuan2021instancerefer} &  75.7 & 64.7 & 29.4 & 23.0 & 38.4 & 31.1 \\
3DVG-Trans~\citep{zhao20213dvg} & 81.9 & 60.6 & 39.3 & 28.4 & 47.6 & 34.7 \\
MVT~\citep{chen2021mvt} & 77.7 & 66.4 & 31.9 & 25.3 & 40.8 & 33.3 \\
3D-SPS~\citep{luo20223d} & 84.1 & 66.7 & 40.3 & 29.8 & 48.8 & 37.0 \\
ViL3DRel~\citep{chen2022language} & 81.6 & 68.6 & 40.3 & 30.7 & 47.9 & 37.7 \\
3DJCG~\citep{cai20223djcg} & 83.5 & 64.3 & 41.4 & 30.8 & 49.6 & 37.3 \\
D3Net~\citep{chen2021d3net} & – & 72.0 & – & 30.1 & – & 37.9 \\
BUTD-DETR~\citep{jain2022bottom} &  84.2 & 66.3 & 46.6 & 35.1 & 52.2 & 39.8 \\
HAM~\citep{chen2022ham} & 79.2 & 67.9 & 41.5 & 34.0 & 48.8 & 40.6 \\
3DRP-Net~\citep{wang20233drp} & 83.1 & 67.7 & 42.1 & 32.0 & 50.1 & 38.9 \\
3D-VLP~\citep{jin2023context} & 84.2 & 64.6 & 43.5 & 33.4 & 51.4 & 39.5 \\
EDA~\citep{wu2023eda} & 85.8 & 68.6 & 49.1 & 37.6 & 54.6 & 42.3 \\
M3DRef-CLIP~\citep{zhang2023multi3drefer} & 85.3 & 77.2 & 43.8 & 36.8 & 51.9 & 44.7 \\
3D-VisTA~\citep{zhu20233d} & 81.6 & 75.1 & 43.7 & 39.1 & 50.6 & 45.8 \\
ConcreteNet~\citep{unal2024four} & 86.4 & 82.1 & 42.4 & 38.4 & 50.6 & 46.5 \\
\midrule
 \textbf{\textit{3D LLMs}}\\
Chat-Scene~\citep{huang2024chat} & 89.6 & 82.5 & 47.8 & 42.9 & 55.5 & 50.2 \\
Video-3D-LLM~\citep{zheng2024video} & 88.0 & 78.3 & 50.9 & 45.3 & 58.1 & 51.7 \\
Ross3D~\citep{wang2025ross3d} & 87.2 & 77.4 & 54.8 & 48.9 & 61.1 & 54.4 \\
GPT4Scene$^{\star}$~\citep{qi2025gpt4scene} & 65.5 & 61.2 & 34.8 & 31.1 & 40.5 & 36.7 \\
GPT4Scene-HD~\citep{qi2025gpt4scene} & 77.5 & 71.9 & 44.9 & 40.6 & 50.9 & 46.4 \\
GPT4Scene-HDM~\citep{qi2025gpt4scene} & {\color{myred}\textbf{90.3}} & {\color{myred}\textbf{83.7}} & {\color{myblue}\textbf{56.4}} & {\color{myblue}\textbf{50.9}} & {\color{myblue}\textbf{62.6}} & {\color{myblue}\textbf{57.0}} \\
    \midrule
    Ours  &  {\color{myblue}\textbf{89.6}} & {\color{myblue}\textbf{83.5}} & {\color{myred}\textbf{56.7}} & {\color{myred}\textbf{51.5}} &  {\color{myred}\textbf{62.7}} & {\color{myred}\textbf{57.4}} \\ 
      \bottomrule
    \end{tabular}
    }
    \label{tab:appendix-scanrefer-detail}
\end{table}

Table~\ref{tab:appendix-multi3drefer-detail} illustrates the comprehensive evaluation results for 3D visual grounding performance on the Multi3DRef~\citep{zhang2023multi3drefer} across five distinct scenarios: Zero Target without Distractors (ZT w/o D), Zero Target with Distractors (ZT w/ D), Single Target without Distractors (ST w/o D), Single Target with Distractors (ST w/ D), and Multi-Target (MT). Performance is assessed through F1 scores at IoU thresholds of 0.25 and 0.5, emphasizing precision in object localization under varying complexity and distractor presence conditions.
The proposed approach demonstrates competitive performance across multiple scenarios, achieving notable results especially in scenarios involving distractors. For instance, our method achieves the highest F1@0.25 (60.0) and F1@0.5 (55.1) scores in the challenging Single Target with Distractors (ST w/ D) scenario, surpassing previous strong models such as GPT4Scene-HDM~\citep{qi2025gpt4scene}. Similarly, in the comprehensive evaluation across all scenarios (denoted ``ALL''), our method attains leading performance (F1@0.25: 65.1, F1@0.5: 60.5), indicating its broad effectiveness in diverse grounding contexts.
Task-specific methods, such as M3DRef-CLIP~\citep{zhang2023multi3drefer} and 3DICG (Grounding)~\citep{cai20223djcg}, exhibit strong performance in simpler settings (e.g., ZT w/o D and ST w/o D), though their results show noticeable declines when encountering scenarios with distractors or multiple targets. 
In contrast, the proposed approach demonstrates enhanced robustness and flexibility in addressing increased task complexity. 
This observation suggests that explicitly modeling multi-modal complexity and integrating Mixture-of-Experts (MoE) module within LLM frameworks may positively influence grounding performance.

\begin{table}[h!]
    \centering
    \caption{Full evaluation results of 3D visual grounding on Multi3DRef~\citep{zhang2023multi3drefer}.    
    Performance is assessed across five scenarios: Zero Target without Distractors (ZT w/o D), where no object matches the referring expression and no distractors exist; Zero Target with Distractors (ZT w/ D), where no object matches but distractors are present; Single Target without Distractors (ST w/o D), referring to a single, uniquely identifiable target object without distractors; Single Target with Distractors (ST w/ D), a single target object with multiple distractors present; and Multi-Target (MT), where multiple objects match the referring expression simultaneously. Metrics reported include the F1 score (F1) at IoU thresholds of 0.25 and 0.5 (F1@0.25, F1@0.5), reflecting localization precision and recall. ``ALL'' aggregates results across all five scenarios. $\star$ indicates that high-resolution settings are not used. We highlight the best performance in {\color{myred}\textbf{red}} and the second-best in {\color{myblue}\textbf{blue}}.} 
    \resizebox{\textwidth}{!}{
    \begin{tabular}{ccccccccccc}
    \toprule
     \multirow{2}{*}{\textbf{Method}} &  \textbf{ZT w/o D} & \textbf{ZT w/ D}  & \multicolumn{2}{c}{\textbf{ST w/o D}}   & \multicolumn{2}{c}{\textbf{ST w/ D}} & \multicolumn{2}{c}{\textbf{MT}}  & \multicolumn{2}{c}{\textbf{ALL}}  \\
     \cmidrule(lr){2-2} \cmidrule(lr){3-3} \cmidrule(lr){4-5} \cmidrule(lr){6-7}  \cmidrule(lr){8-9}  \cmidrule(lr){10-11} 
       & F1 & F1 & F1@0.25 & F1@0.5 & F1@0.25 & F1@0.5 & F1@0.25 & F1@0.5 & F1@0.25 & F1@0.5 \\
     \midrule
     \textbf{Task-Specific Model} \\
3DVG-Trans~\citep{zhao20213dvg} & 87.1 & 45.8 & – & 27.5 &  – & 16.7 & – & 26.5 & – & 25.5 \\
D3Net (Grounding)~\citep{chen2021d3net} &  81.6 & 32.5 & – & 38.6 & – & 23.3 & – & 35.0 &  – &  32.2 \\
3DJCG (Grounding)~\citep{cai20223djcg} & 94.1 & 66.9 & – & 26.0 & – & 16.7 & – & 26.2 & – & 26.6 \\
M3DRef-CLIP~\citep{zhang2023multi3drefer} & 81.8 & 39.4 & 53.5 & 47.8 & 34.6 & 30.6 & 43.6 &  37.9 &  42.8 &  38.4 \\
\midrule
   \textbf{3D LLMs} \\
Chat-Scene [39] & 90.3 & 62.6 & 82.9 & 75.9 & 49.1 & 44.5 & 45.7 & 41.1 & 57.1 & 52.4 \\
GPT4Scene$^{\star}$~\citep{qi2025gpt4scene} & 85.2 & 61.4 & 60.1 & 55.1 & 37.7 & 34.4 & 39.4 & 36.3 & 45.4 & 42.1 \\
GPT4Scene-HD~\citep{qi2025gpt4scene} & 93.6 & 81.8 & 72.5&  66.2 & 46.6 &  42.9 &  41.8 & 38.9 & 53.7 &  50.0 \\
GPT4Scene-HDM~\citep{qi2025gpt4scene} &  {\color{myred}\textbf{97.4}} & {\color{myblue}\textbf{84.4}} &  {\color{myred}\textbf{85.0}} & {\color{myred}\textbf{77.7}} &  {\color{myblue}\textbf{59.9}} & {\color{myred}\textbf{55.1}} & {\color{myblue}\textbf{48.6}} &  {\color{myblue}\textbf{44.6}} &{\color{myblue}\textbf{64.5}} & {\color{myblue}\textbf{59.8}} \\
\midrule
Ours &   {\color{myblue}\textbf{96.8}} & {\color{myred}\textbf{84.7}} & {\color{myblue}\textbf{84.9}} & {\color{myblue}\textbf{77.3}} & {\color{myred}\textbf{60.0}} & {\color{myred}\textbf{55.1}} & {\color{myred}\textbf{51.4}} & {\color{myred}\textbf{47.7}} &   {\color{myred}\textbf{65.1}} &  {\color{myred}\textbf{60.5}} \\
     
      \bottomrule
    \end{tabular}
    }
    \label{tab:appendix-multi3drefer-detail}
\end{table}

\textbf{3D Dense Captioning.}
Table~\ref{tab:appendix-scan2cap} presents the evaluation results of 3D dense captioning on the Scan2Cap~\citep{chen2021scan2cap} benchmark, comparing our model against several state-of-the-art methods. Performance is measured using widely adopted captioning metrics—BLEU-4, METEOR, ROUGE, and CIDEr—at IoU thresholds of 0.25 and 0.5, indicating the quality and spatial accuracy of generated captions.

Our method achieves superior performance compared to other advanced approaches, including both task-specific models and recent 3D LLMs. Specifically, at IoU=0.25, our model attains the highest BLEU-4 (44.4), METEOR (29.9), and CIDEr (89.9) scores, indicating strong fluency, semantic alignment, and relevance of the captions. 
At a stricter threshold of IoU=0.5, our model also demonstrates leading performance with top results in BLEU-4 (41.1) and CIDEr (85.2), highlighting the method’s robustness in precise localization conditions. 
These findings underscore the advantage of integrating the Mixture-of-Experts architecture into multimodal language models, improving caption generation in complex 3D scenarios.

\begin{table}[h!]
    \centering
    \caption{Evaluation results of 3D dense captioning on Scan2Cap~\citep{chen2021scan2cap}. BLEU-4, METEOR, ROUGE and CIDEr denote text similarity scores between the predicted answer and the ground-truth answer. Metrics are computed under IoU thresholds of 0.25 and 0.5 between the predicted and reference bounding boxes. $\star$ indicates that high-resolution settings are not used. We highlight the best performance in {\color{myred}\textbf{red}} and the second-best in {\color{myblue}\textbf{blue}}.}
    \resizebox{\textwidth}{!}{
    \begin{tabular}{ccccccccc}
    \toprule
     \multirow{2}{*}{\textbf{Method}} &  \multicolumn{4}{c}{\textbf{Scan2Cap} (IoU@0.25)}   & \multicolumn{4}{c}{\textbf{Scan2Cap} (IoU@0.5)} \\
     \cmidrule(lr){2-5} \cmidrule(lr){6-9}  
       & BLEU-4 & METEOR & ROUGE & CIDEr & BLEU-4 & METEOR & ROUGE & CIDEr \\
     \midrule
     \textbf{\textit{Task-specific Models}} \\
    Scan2Cap~\citep{chen2021scan2cap} & 34.2 & 26.3  & 55.3 & 56.8 & 22.4 & 21.4 & 43.5 & 35.2 \\
    3DJCG~\citep{cai20223djcg} & 40.2 & 27.7 & 59.2 & 64.7 &  31.5 & 24.3 & 51.8 &  47.7 \\
    3D-VLP\citep{jin2023context} & 41.0 & 28.1 & 59.7 & 70.7 & 32.3 &  24.8 & 51.5  & 54.9 \\
    3D-VisTA~\citep{zhu20233d} & 36.5 & 28.4 & 57.6 &  71.0 & 34.0 &  26.8 &  54.3 & 61.6 \\
    Vote2Cap-DETR~\citep{chen2023end} & 39.3  & 28.3 & 59.3 & 71.5 & 34.5 & 26.2 & 54.4 & 61.8 \\
    X-Trans2Cap~\citep{yuan2022x} & 35.7 & 26.6 & 54.7 & 61.8 & 25.1 & 22.5 & 45.3 & 43.9 \\
    
      \midrule
      \textbf{\textit{3D LLMs}}\\
    LEO~\citep{huang2024embodied} & - & - & - & - & 38.2 & 27.9 & 58.1  & 72.4\\
    LL3DA~\citep{chen2024ll3da} &  41.4 & 27.8 &  59.5 & 74.2 & 36.8 & 26.0 & 55.1 & 65.2 \\
    Chat-Scene~\citep{huang2024chat} & 38.2 & 29.0 & 60.6 & 81.9 & 36.3 & 28.0 & 58.1 & 77.1 \\
    LLaVA-3D~\citep{zhu2024llava} & - & - & - & - & 41.1 & 30.2 & 63.4 & 79.2\\
    Ross3D~\citep{wang2025ross3d} & - & - & - & - & 43.4 & 30.3 & 66.9 & 81.3 \\
    GPT4Scene$^{\star}$~\citep{qi2025gpt4scene} & 36.3 & 26.5 & 57.6 & 63.8 & 34.2 & 25.6 & 55.2 & 60.6\\
    GPT4Scene-HD~\citep{qi2025gpt4scene} & 40.4 & 28.3 & 60.2 & 79.1 & 37.9 & 27.3 & 57.7 & 74.4 \\
    GPT4Scene-HDM~\citep{qi2025gpt4scene} &  {\color{myblue}\textbf{43.1}} & {\color{myblue}\textbf{29.3}} & {\color{myred}\textbf{61.9}} & {\color{myred}\textbf{91.7}} & {\color{myblue}\textbf{40.6}} & {\color{myblue}\textbf{28.2}} & {\color{myred}\textbf{59.3}} & {\color{myred}\textbf{86.3}} \\
    \midrule
      Ours & {\color{myred}\textbf{44.4}} & {\color{myred}\textbf{29.9}} & {\color{myblue}\textbf{60.7}} & {\color{myblue}\textbf{89.9}} &  {\color{myred}\textbf{41.1}} & {\color{myred}\textbf{29.9}} & {\color{myred}\textbf{59.3}} &  {\color{myblue}\textbf{85.2}} \\

      \bottomrule
    \end{tabular}
    }
    \label{tab:appendix-scan2cap}
\end{table}

\section{Additional Results for the MoE Module}\label{appendix:moe-results}

\subsection{Visualization Results of MoE}\label{appendix:moe-vidual}
\begin{figure}[h!]
    \centering
    \includegraphics[width=1.0\linewidth]{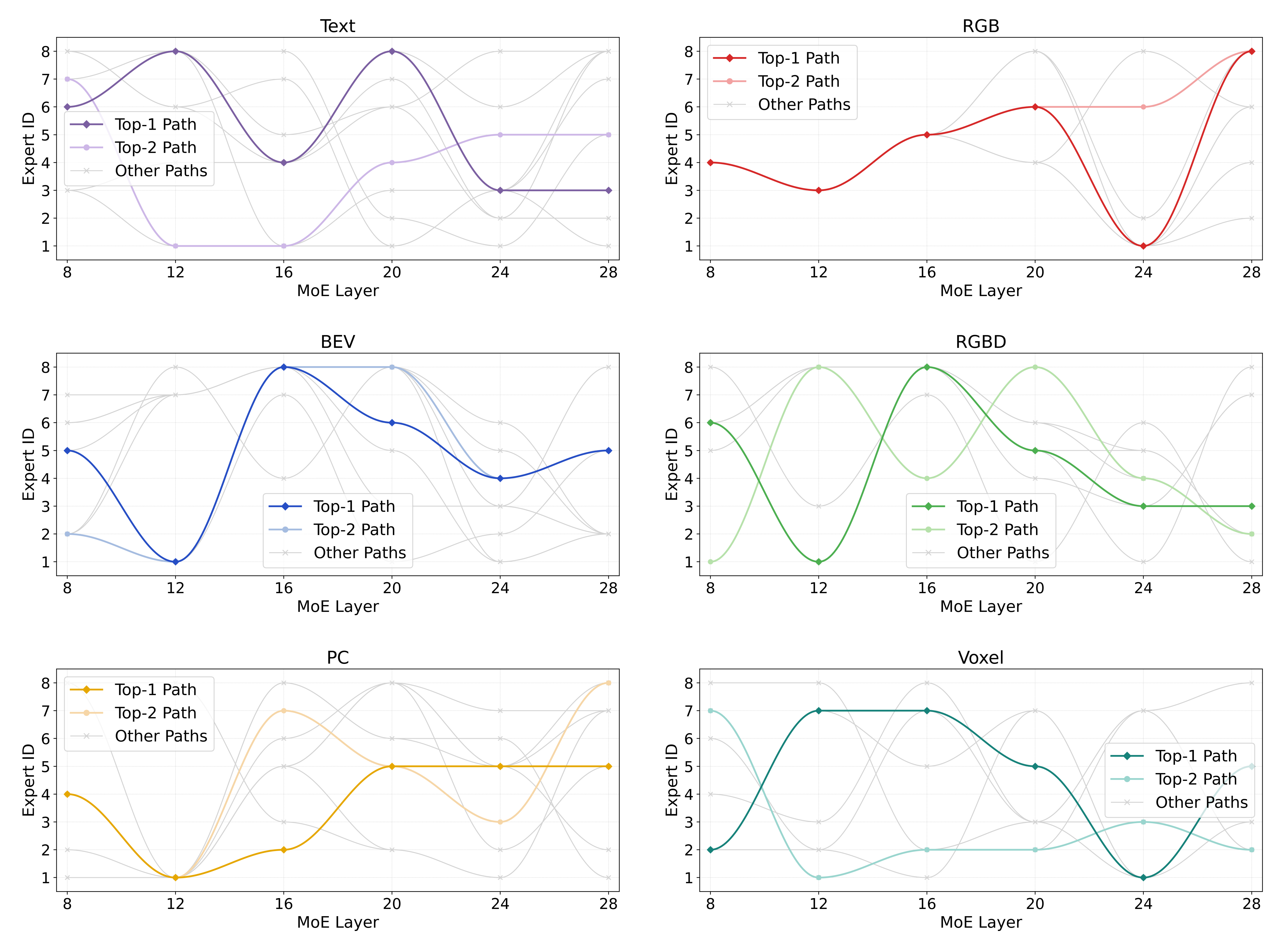}
    \caption{Top-10 activated routing pathways for different modalities, highlighting dynamic and specialized expert activation. Colored curves illustrate the top-1 and top-2 routing paths for each modality, while gray curves represent the remaining eight pathways. }
    \label{fig:route-path-top10}
\end{figure}

\textbf{Top-10 Activated Routing Pathways.}
As illustrated in Fig.~\ref{fig:route-path-top10}, we visualize the top-10 activated routing pathways across different modalities (Text, RGB, BEV, RGBD, PC, and Voxel) through multiple Mixture-of-Experts (MoE) layers. Each modality is represented in a distinct color, with the most prominent paths (Top-1 and Top-2) highlighted, while the other pathways are depicted in gray to emphasize relative activation strengths. The visualization reveals dynamic and specialized expert activations that vary across layers, highlighting the model’s adaptive routing mechanism. For instance, point cloud (PC) modality prominently engages expert $\mathcal{E}_5$ at deeper layers (layers 20, 24, and 28), whereas RGB modality dynamically shifts its primary expert from expert $\mathcal{E}_4$ at layer 8 to expert $\mathcal{E}_1$ at layer 24. This behavior underscores the importance of employing MoE architectures to dynamically allocate modality-specific information to the most suitable experts at different representation depths, thereby enhancing overall model performance.

\begin{figure}[h!]
    \centering
    \includegraphics[width=1.0\linewidth]{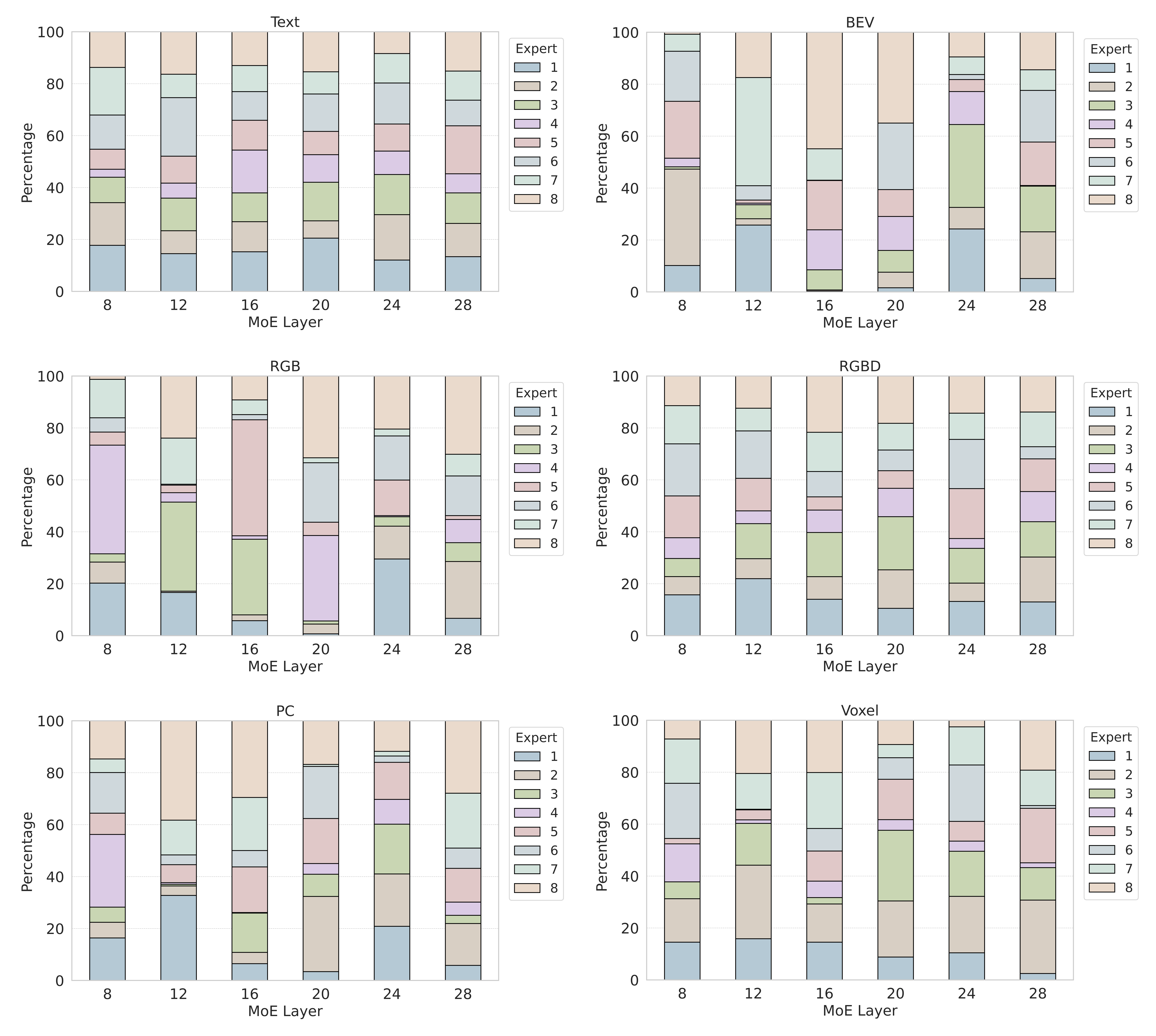}
    \caption{Expert assignment distribution for each modality across MoE layers in Uni3D-MoE, highlighting modality-specific expert selection dynamics.}
    \label{fig:modality}
\end{figure}

\textbf{Expert Assignment Distribution for Each Modality.}
Fig.~\ref{fig:modality} illustrates the varying patterns of expert assignment for each modality (Text, BEV, RGB, RGBD, PC, and Voxel) across different MoE layers (layers 8, 12, 16, 20, 24, and 28) within the Uni3D-MoE model. Observations suggest potential modality-specific preferences and evolving trends in expert usage as model depth increases. 
For example, the RGB modality distinctly varies its expert selection patterns across layers: prominently activating expert $\mathcal{E}_4$ at layer 8, experts $\mathcal{E}_3$ and $\mathcal{E}_5$ at layers 12 and 16, and shifting focus towards experts $\mathcal{E}_6$ and $\mathcal{E}_8$ in deeper layers (layers 20, 24, and 28). This progression may reflect changing requirements in visual feature extraction as information abstraction deepens.
Similarly, the BEV modality noticeably engages experts $\mathcal{E}_2$, $\mathcal{E}_7$ and $\mathcal{E}_8$ frequently at intermediate layers but seems to diversify at deeper layers, possibly due to increasing complexity in spatial reasoning tasks. 
The Voxel modality frequently utilizes experts $\mathcal{E}_2$, $\mathcal{E}_3$ and $\mathcal{E}_7$, which might indicate specific geometric feature processing demands. 
The PC modality prominently selects experts $\mathcal{E}_1$ and $\mathcal{E}_8$ in the initial layers (layers 8, 12, and 16), while in deeper layers (20, 24, and 28) it shifts predominantly towards experts $\mathcal{E}_2$ and $\mathcal{E}_8$. This pattern might indicate evolving requirements in geometric or structural feature extraction as the model processes point cloud data at different abstraction levels. 
The Text and RGBD modalities exhibit relatively uniform expert distribution patterns across layers, suggesting stable and balanced processing demands possibly related to semantic and visual-depth integration tasks. 
These patterns collectively highlight Uni3D-MoE’s capability to potentially adapt routing strategies for different modalities, thus possibly enhancing multimodal representation effectiveness and task-specific performance.

\subsection{Ablation Results of MoE}\label{appendix:moe-abl}

\textbf{Ablation on MoE Layer Placement.}
Table~\ref{tab:ab-moe-layer} presents the ablation results for integrating the Mixture-of-Experts (MoE) module at different layers in the ScanQA~\citep{azuma2022scanqa} benchmark. We observe a clear performance improvement across all evaluation metrics when employing the MoE module at deeper layers. Specifically, incorporating MoE layers at depths [8,12,16,20,24,28] achieves the best results, significantly enhancing EM@1 accuracy from 27.3\% (without MoE) to 30.8\%. Similar improvements are evident across other metrics, such as CIDEr, which rises notably from 88.4 to 97.6. These results highlight that deeper integration of the MoE mechanism facilitates richer feature extraction, thereby enhancing the model's ability to accurately answer complex questions.

\begin{table}[t!]
    \centering
    \small
    \caption{Ablation results of MoE layers on ScanQA\citep{azuma2022scanqa}.}
    \resizebox{\textwidth}{!}{
    \begin{tabular}{lllllllllll}
        \toprule
        \textbf{Method} & \textbf{MoE layers} & \textbf{EM@1} & \textbf{EM-R@1} & \textbf{F1} &  \textbf{BLEU-1} & \textbf{BLEU-4} & \textbf{METEOR} & \textbf{ROUGE} & \textbf{CIDEr} \\
        \midrule
         w/o MoE & - & 27.3 & 45.1 & 45.5 & 41.9 & 13.9 & 17.1 & 43.8 &  88.4  \\ 
         
         w/ MoE & [0,2,4,6,8,10] & 27.7 & 45.3 & 46.2 & 41.5 & 14.1 & 17.6 & 44.0 & 90.9  \\
         w/ MoE & [0,4,8,12,16,20] & 28.7 & 47.1 & 47.8 & 42.2 & 15.7 & 17.9 & 46.0 & 94.3  \\
         \rowcolor{red!5}
         w/  MoE & [8,12,16,20,24,28] & 30.8 & 49.0 & 48.8 & 43.7 & 17.5 & 19.0 & 47.1 & 97.6 \\
         \bottomrule
    \end{tabular}
    }
    \label{tab:ab-moe-layer}
\end{table}

\textbf{MoE Ablation Across Benchmarks.}
Tables~\ref{tab:appendix-abl-sqa3d}-\ref{tab:appendix-abl-scan2cap} present additional ablation results of the Mixture-of-Experts (MoE) module on different 3D scene understanding tasks.
Specifically, on the SQA3D~\citep{ma2022sqa3d} benchmark, integrating MoE yields clear improvements, achieving better exact-match accuracy (EM@1: 57.2 vs. 54.6) and significantly higher text similarity scores (e.g., CIDEr: 147.8 vs. 136.0). 
For visual grounding tasks on ScanRefer~\citep{chen2020scanrefer} and Multi3DRefer~\citep{zhang2023multi3drefer}, the MoE-equipped GPT4Scene backbone consistently shows performance gains (ScanRefer Acc@0.5: 57.4 vs. 57.0; Multi3DRefer F1@0.5: 60.5 vs. 59.8). In the 3D dense captioning scenario (Scan2Cap~\citep{chen2021scan2cap}), the MoE integration brings mixed results, slightly improving BLEU-4 and METEOR at IoU@0.25 but slightly decreasing CIDEr scores. 
Overall, the introduction of MoE demonstrates consistent, albeit varied, improvements across multiple 3D scene understanding tasks.

\begin{table}[h!]
    \centering
    \caption{Ablation results of the MoE module on SQA3D test set~\citep{ma2022sqa3d} for 3D question answering. EM@1 refers to the top-1 exact match accuracy; BLEU-1, BLEU-4, METEOR, and CIDEr denote text similarity scores between the predicted and ground-truth answer.}
    \resizebox{\textwidth}{!}{
    \begin{tabular}{cccccccc}
    \toprule
     \textbf{Method} & EM@1 & EM-R@1 & BLEU-1 & BLEU-4 & METEOR & ROUGE & CIDEr \\
     \midrule
   w/o MoE & 54.6 & 55.7 & 50.4 & 39.6 & 35.3 & 52.8 &  136.0 \\
    \rowcolor{red!5}
    w/ MoE  & 57.2 & 59.8 & 54.9 & 43.5 & 38.3 & 57.9 & 147.8 \\
      \bottomrule
    \end{tabular}
    }
    \label{tab:appendix-abl-sqa3d}
\end{table}

\begin{table}[h!] 
    \centering
    \caption{Ablation results of the MoE module on ScanRefer~\citep{chen2020scanrefer} Multi3DRefer~\citep{zhang2023multi3drefer} for 3D visual grounding.}
    \begin{tabular}{ccccccccc}
    \toprule
     \multirow{2}{*}{\textbf{Method}} &  \multicolumn{2}{c}{\textbf{ScanRefer}} &  \multicolumn{2}{c}{\textbf{Multi3DRefer}}    \\
     \cmidrule(lr){2-3} \cmidrule(lr){4-5} 
       & Acc@0.25 & Acc@0.5 & F1@0.25 & F1@0.5  \\ 
    \midrule
    w/o MoE & 62.6 & 57.0 & 64.5 & 59.8 \\
    \rowcolor{red!5}
    w/ MoE & 62.7 & 57.4 & 65.1 & 60.5  \\
      \bottomrule
    \end{tabular}
    \label{tab:appendix-abl-scanrefer}
\end{table}

\begin{table}[h!]
    \centering
    \caption{Ablation results of the MoE module on Scan2Cap~\citep{chen2021scan2cap} for 3D dense captioning. BLEU-4, METEOR, and CIDEr denote text similarity scores between the predicted answer and the ground-truth answer. Metrics are computed under IoU thresholds of 0.25 and 0.5 between the predicted and reference bounding boxes. $\star$ indicates that high-resolution settings are not used.}
    \resizebox{\textwidth}{!}{
    \begin{tabular}{ccccccccccc}
    \toprule
    \multirow{2}{*}{\textbf{Method}} &  \multicolumn{4}{c}{\textbf{Scan2Cap} (IoU@0.25)}   & \multicolumn{4}{c}{\textbf{Scan2Cap} (IoU@0.5)} \\
     \cmidrule(lr){2-5} \cmidrule(lr){6-9}  
       & BLEU-4 & METEOR & ROUGE & CIDEr & BLEU-4 & METEOR & ROUGE & CIDEr \\
     \midrule
    w/o MoE & 43.1 & 29.3 & 61.9 & 91.7 & 40.6 & 28.2 & 59.3 & 86.3 \\
        \rowcolor{red!5}
   w/ MoE & 44.4 & 29.9 & 60.7 & 89.9 & 41.1 & 29.9 & 59.3 & 85.2 \\
      \bottomrule
    \end{tabular}
    }
    \label{tab:appendix-abl-scan2cap}
\end{table}

\section{Limitations and Broader Impacts}\label{appendix:limitations+bi}
\textbf{Limitations.}
Despite achieving promising results across various tasks, Uni3D-MoE still exhibits several limitations. 
First, the token budget constraints of large language models necessitate strict control over modality-specific inputs. 
To this end, multi-view images are selected using the Maximum Voxel Coverage Sampling (MVCS) algorithm. 
While effective, this method may overlook critical viewpoints, leading to an incomplete spatial context.
Similarly, point clouds are downsampled using Farthest Point Sampling (FPS), reducing point density and limiting the representation of fine-grained object details—particularly for small-scale structures. 
These input reductions might degrade model performance, especially when other modalities fail to provide sufficient complementary information.
Second, the model’s effectiveness is partially constrained by the quality of the training dataset. Blurry multi-view images and annotation inaccuracies introduce noise and ambiguity, which can hinder performance in tasks that demand precise spatial understanding and accurate object localization.

\textbf{Broader Impacts.}
This paper aims to enhance the 3D perception capabilities of VLM. The proposed Uni3D-MoE has potential applications in human-computer interaction and autonomous robotics. It can help embodied agents better understand the environment and perform complex tasks. While there are potential concerns about misuse, such as applications in military robotics, we believe the benefits of our approach significantly outweigh the minimal risks.

\section{Data Details}\label{appendix:data}
\begin{figure}[h!]
    \centering
    \includegraphics[width=0.8\linewidth]{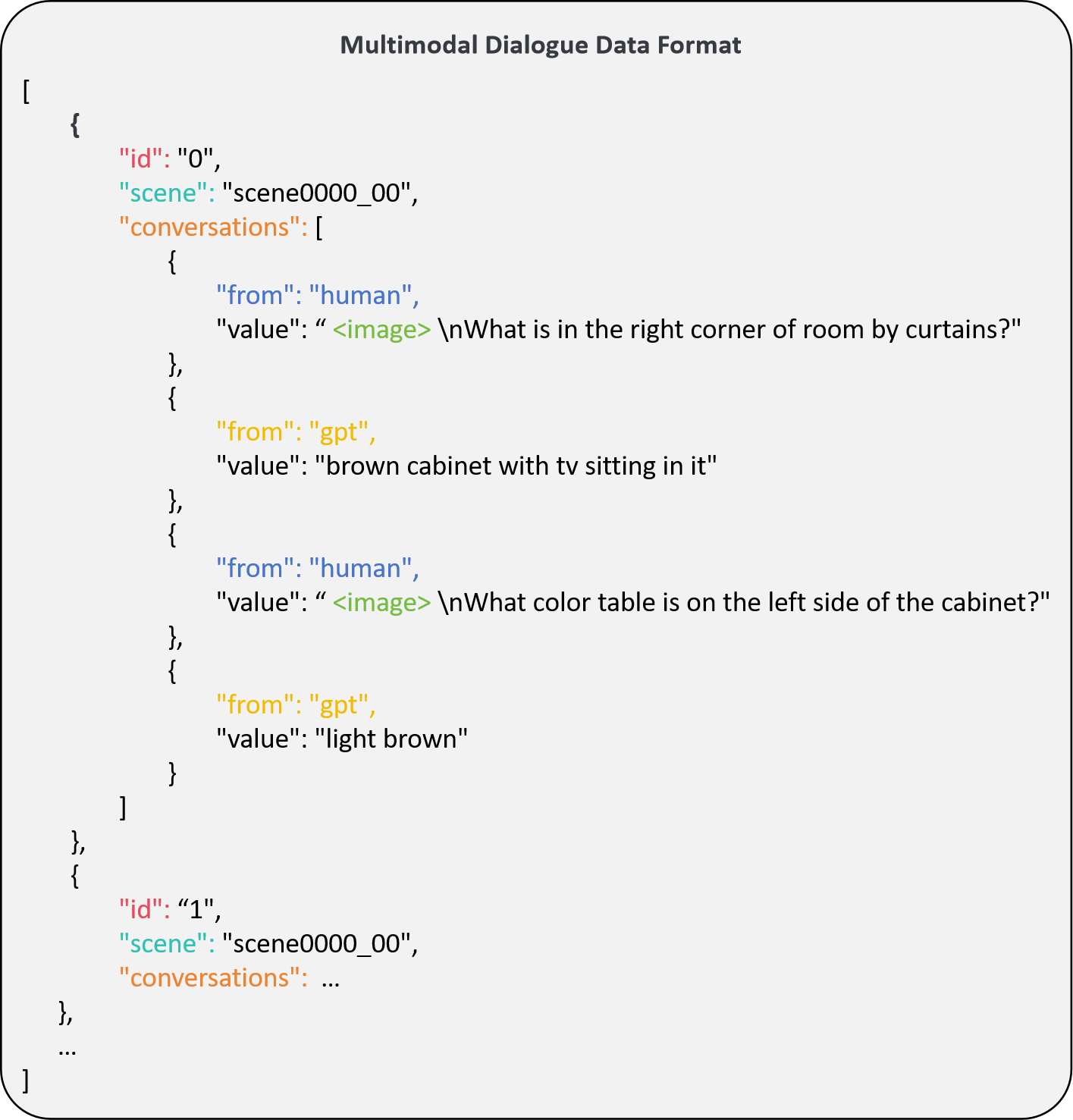}
    \caption{Multimodal dialogue data format.}
    \label{fig:data-format}
\end{figure}

\begin{figure}[h!]
    \centering
    \includegraphics[width=0.8\linewidth]{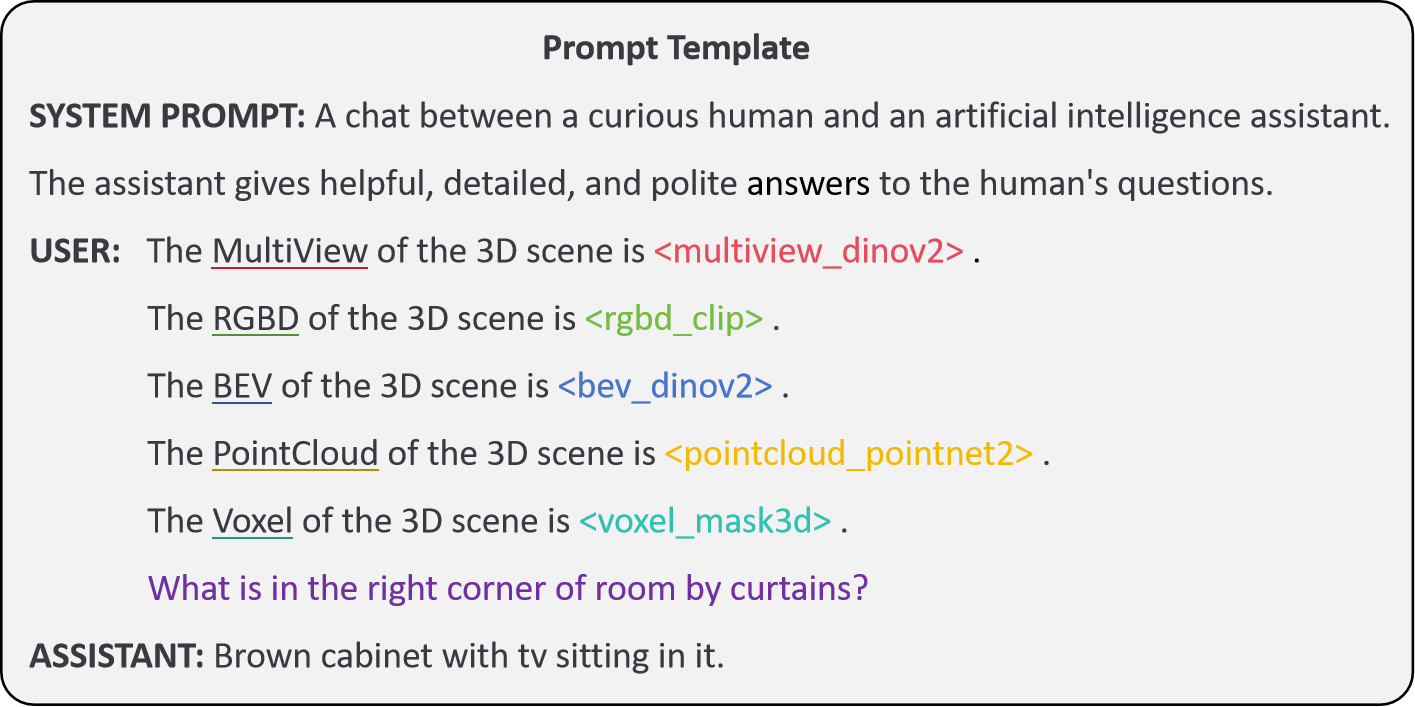}
    \caption{Prompt template.}
    \label{fig:prompt-template}
\end{figure}

\begin{figure}[h!]
    \centering
    \includegraphics[width=0.8\linewidth]{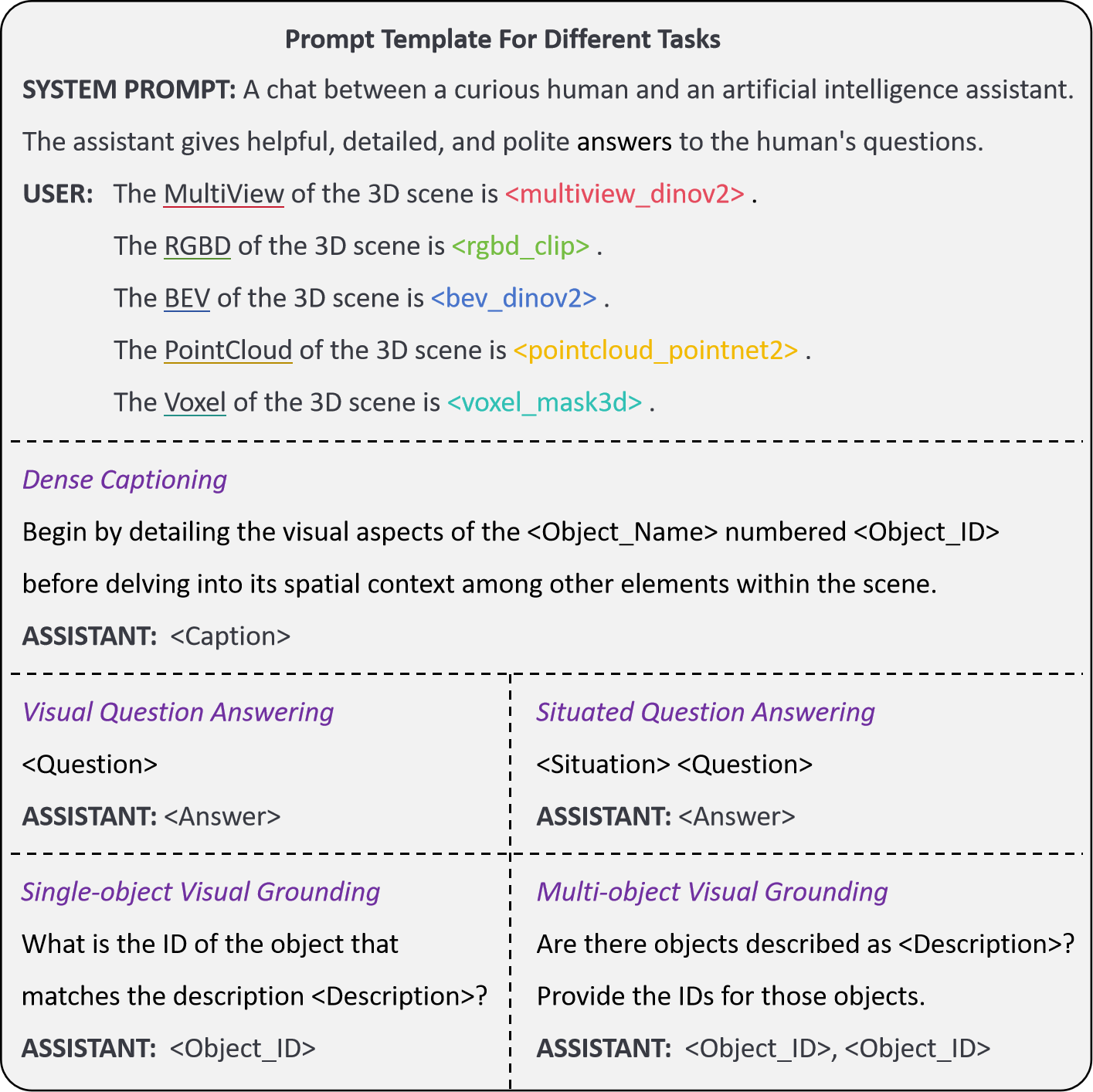}
    \caption{Prompt Template for different tasks.}
    \label{fig:prompt-template-task}
\end{figure}

Fig.s~\ref{fig:data-format}, \ref{fig:prompt-template}, and \ref{fig:prompt-template-task} illustrate the data organization and prompting approach used for multimodal dialogue tasks.

Fig.~\ref{fig:data-format} illustrates the structure of our multimodal dialogue data format. Each dialogue instance is grounded in a specific 3D scene, denoted by the ``scene'' field. The ``conversations'' field contains a sequence of interactions that revolve around this scene, capturing the exchange between the human user and the model. Each turn is marked by a ``from'' field (``human'' or ``gpt'') and a corresponding ``value'', which includes multimodal placeholders ``<image>'' in the first round. An ``id'' is also assigned to each dialogue instance for indexing purposes.

Fig.s~\ref{fig:prompt-template} and~\ref{fig:prompt-template-task} illustrate the prompt design used in our multimodal dialogue system. Each prompt begins with a system message that sets the context for a conversation between a human and an AI assistant. The user input includes a list of available 3D modality features—such as MultiView, RGBD, BEV, PointCloud, and Voxel—each referenced by a corresponding placeholder token (e.g., \texttt{<multiview\_dinov2>}). These tokens serve as modality-specific representations rather than raw input data.

Fig.~\ref{fig:prompt-template-task} further categorizes prompt formats according to different task types. For dense captioning, the assistant is prompted to describe the appearance and spatial context of a specific object based on its name and ID. For question answering tasks, the user issues natural language queries, optionally with contextual grounding. Visual grounding tasks ask the assistant to return object identifiers that match given textual descriptions. 
The object IDs used in these tasks are typically generated by an instance segmentation model such as Mask3D~\citep{schult2023mask3d}, ensuring consistent identification across the 3D scene understanding dataset.

\section{Model Details}\label{appendix:model-details}
\begin{algorithm}[htbp] 
  \caption{Maximum Voxel Coverage Sampling(MVCS) with Voxel Pruning and View Refinement}
  \label{appendix:alg-enhanced-mvcs}
  \begin{algorithmic}[1]
    \Require Scene voxel set $V$, camera params $\{C_k\}$, budget $K$, distance limit $d_{\max}$
    \Ensure Selected view set $S$
    \Function{Sampling}{$V$, $\{C_k\}$, $K$, $d_{\max}$}
    \State $V_{\mathrm{scene}} \gets \{\,v\in V \mid v.\text{type} \in \{\text{floor},\text{ceiling},\text{wall}\} \}$ \Comment{1. Voxel pruning}
    \For{each view $f_k$}
      \State $V_k \gets \{\,v\in V_{\mathrm{scene}} \mid \text{visible}(\text{proj}(v, C_k))\text{ and } \|X_v - X_{C_k}\| \le d_{\max}\,\}$
    \EndFor
    \State 
    \State $S \gets \varnothing$, \quad $U \gets \varnothing$  \Comment{2. Perform greedy selection based on marginal coverage}
    \While{$|S| < K$}
      \For{each view $f_k \notin S$}
        \State $g_k \gets |V_k \setminus U|$
      \EndFor
      \State $f^* \gets \arg\max_k g_k$
      \State $S \gets S \cup \{f^*\}$,\quad $U \gets U \cup V_{f^*}$
    \EndWhile

    \State 
\For{each index $i$ such that $f_i \in S$}\Comment{3. Selected clear views}
  \State $N \gets \{\, f_j \mid j \in [\max(0, i{-}2), \min(i{+}2, n{-}1)] \,\}$
  \For{each $f_j \in N$}
    \State $s_j \gets \mathrm{Var}(\nabla^2 I_{f_j})$
  \EndFor
  \State $f_{\text{best}} \gets \arg\max_{f_j \in N} s_j$
  \State $f_i \gets f_{\text{best}}$
\EndFor

    \State \Return $S$
    \EndFunction
  \end{algorithmic}
\end{algorithm}

\subsection{Modality-specific Encoders}\label{appendix:model-encoder}

\textbf{Multi-view RGB Encoder.} Given an RGB-D video $\mV \in \R^{V_v \times H \times W \times 3}$, we adopt Maximum Voxel Coverage Sampling (MVCS) (shown in Alg.~\ref{appendix:alg-enhanced-mvcs}) to select informative keyframes as Multi-view RGB Images $\mI \in \R^{V \times H \times W \times 3}$, where $V_v$ is the video length, $H$ and $W$ are the image height and width, and $V$ is the predefined number of views. In our implementation, we set $V$=24. Compared to previous methods~\citep{zheng2024video}, our algorithm achieves 100× speed-up in computing coverage by using camera poses instead of depth maps. To focus on task-relevant objects, the algorithm removes low-contribution scene segments (e.g., floor), and voxels located too far from the camera are considered not covered. Finally, for each maximum coverage frame, the algorithm selects its clearest neighbor as the keyframe using Laplacian variance. We use a pre-trained 2D encoder (e.g., DINOv2~\citep{oquab2023dinov2}) with a trainable modal projector to convert Multi-view RGB Images $\mI$ into visual tokens $\mF_{rgb} \in \R^{VN_{rgb} \times D_{rgb}}$, where $N_{rgb}$ is the number of tokens per image, and $D_{rgb}$ is the feature dimension.

\textbf{Multi-view Depth Encoder.}
Inspired by~\citep{zhu2024llava}, we leverage depth images $\mD \in \R^{V \times H \times W \times 1}$ and camera parameters to back-project each 2D patch into 3D space, obtaining its corresponding 3D position.
The 2D patch tokens are first extracted from multi-view RGB images using CLIP~\citep{radford2021learning}.
These 3D positions are then encoded into 3D embeddings, which are added to 2D patch tokens to form spatially-aware 3D patches.
To reduce sequence length while preserving spatial context, we apply a 3D-aware pooling and obtain the final RGB-D features $\mF_{rgbd} \in \R^{VN_{rgbd} \times D_{rgbd}}$, where $N_{rgbd}$ and $D_{rgbd}$ represent the token count and feature dimension.

\textbf{BEV Map Encoder.} 
Egocentric images/videos typically lack global scene context, making it difficult for models to understand the overall spatial layout.
To address this, we render the 3D mesh in a bird’s-eye view (BEV) image $\mB \in \R^{H \times W \times 3}$, where $H$ and $W$ are the height and width, respectively.
To enhance object-level understanding, we incorporate instance segmentation into the BEV using numeric labels and colored regions, offering explicit semantic cues.
Then, we use DINOv2~\citep{oquab2023dinov2} to extract BEV features $\mF_{bev} \in \R^{N_{bev} \times D_{bev}}$, where $N_{bev}$ denotes the number of tokens and $D_{bev}$ is the feature dimension.

\textbf{Point Cloud Encoder.}
We apply Farthest Point Sampling (FPS)~\citep{qi2017pointnet++} to the scene-level point cloud to obtain $\mP \in \R^{N \times C}$, where $N$ is the number of sampled points and $C$ includes the 3D coordinates along with additional attributes such as color, normals, and semantic labels.
The sampled points $P$ are then passed through a pre-trained PointNet++~\citep{qi2017pointnet++} backbone to produce point features $\mF_{pc} \in \R^{N_{pc} \times D_{pc}}$, where $N_{pc}$ is the number of point tokens and $D_{pc}$ is the feature dimension. In our implementation, we use farthest point sampling (FPS) to acquire 8,192 sampled points ($N_{pc}=8,192$), where each point contains XYZ coordinates and RGB color attributes ($D_{pc}=6$).

\textbf{Voxel Grid Encoder.}
To extract voxel features, we first voxelize the entire 3D scene and obtain the sparse voxel inputs $\mX \in \R^{M \times C^{\prime}}$. Here, $M$ is the number of non-empty voxels and $C^{\prime}$ is the feature dimension, including sparse tensor coordinates and additional attributes.
The voxelized input $\mX$ is then fed into Mask3D~\citep{schult2023mask3d}, a sparse convolutional U-Net backbone equipped with downsampling and upsampling layers to capture hierarchical context.  
The voxel-level features are then assigned to their corresponding pre-generated segments, followed by segment-wise average pooling to produce high-level representations $\mF_{voxel} \in \R^{N_{voxel} \times D_{voxel}}$.
Here, $N_{voxel}$ and $D_{voxel}$ are the number and dimension of voxel tokens, respectively.

Subsequently, tokens from five modalities are aligned to the text space via respective adapters: 
$\mF_{m}^{\prime} = \text{Adapter}_{m}(\mF_{m}) \in \R^{N_{m} \times D_{txt}}$, where $m \in \{\text{rgb}, \text{rgbd}, \text{bev}, \text{pc}, \text{voxel}\}$ is the modality type and $D_{txt}$ is the target embedding dimension.
Finally, the text prompt feature $\mF_{txt}$, combined with modality-aligned features $\mF_{m}^{\prime}$, composes the unified 3D scene representation:
$\mF_{uni} = \{ \mF_{txt}, \mF_{rgb}^{\prime}, \mF_{rgbd}^{\prime}, \mF_{bev}^{\prime}, \mF_{pc}^{\prime}, \mF_{voxel}^{\prime}\} \in \R^{N_{uni} \times D_{txt}}$, where $N_{uni}$ is the total token number.

\subsection{Modality-specific Adapters.}\label{appendix:model-adapter}
To effectively integrate diverse 3D scene representations into the shared embedding space of the language model, we design modality-specific adapters tailored to the characteristics of each input modality. Each adapter employs a lightweight two-layer MLP projection head to map modality-specific embeddings into the unified 4096-dimensional embedding space required by the language backbone, thereby facilitating efficient multimodal feature alignment and fusion.

\textbf{Multi-view RGB Adapter.}
The multi-view RGB adapter processes concatenated multi-view RGB features, typically aggregating visual details from multiple camera views (e.g., 8 views), resulting in a 12288-dimensional input. It projects these into the common embedding space via:
$
\text{Linear}(12288 \rightarrow 4096) \rightarrow \text{GELU} \rightarrow \text{Linear}(4096 \rightarrow 4096) \rightarrow \text{LayerNorm}.
$
This structure effectively reduces dimensional redundancy from multiple views and normalizes feature distributions for stable integration.

\textbf{RGBD Adapter.}
The RGBD adapter operates on 1024-dimensional spatially-aware RGBD embeddings, leveraging depth-enhanced RGB features. It applies:
$
\text{Linear}(1024 \rightarrow 4096) \rightarrow \text{GELU} \rightarrow \text{Linear}(4096 \rightarrow 4096),
$
to unify depth-aware visual cues with the broader modality embedding space.

\textbf{BEV Adapter.}
The BEV adapter receives BEV-encoded features from \texttt{BEVDinov2Encoder}, which inherently capture spatial structures from a top-down viewpoint in 1536-dimensional embeddings. It aligns them via:
$
\text{Linear}(1536 \rightarrow 4096) \rightarrow \text{GELU} \rightarrow \text{Linear}(4096 \rightarrow 4096) \rightarrow \text{LayerNorm}.
$
This ensures consistent spatial semantic representations across modalities.

\textbf{Point Cloud Adapter.}
The point cloud adapter adapts the sparse geometric features extracted by the \texttt{PointNet2SegEncoder}, which generates a compact 256-dimensional representation from raw point cloud data. It expands and aligns these features using:
$
\text{Linear}(256 \rightarrow 4096) \rightarrow \text{GELU} \rightarrow \text{Linear}(4096 \rightarrow 4096) \rightarrow \text{LayerNorm},
$
preserving critical geometric semantics for downstream tasks.

\textbf{Voxel Adapter.}
The voxel adapter uniquely incorporates five parallel linear branches tailored to voxel features of varying dimensionalities (256, 128, or 96), accommodating voxel grids captured at multiple spatial resolutions. Each branch independently projects features via:
$
\text{Linear} \rightarrow \text{LayerNorm} \rightarrow \text{GELU} \rightarrow \text{Dropout} \rightarrow \text{Linear} \rightarrow \text{LayerNorm},
$
producing uniform 4096-dimensional embeddings to robustly represent volumetric structural information across scales.

By individually tailoring adapter structures to the intrinsic properties of each input modality, these designs collectively ensure robust alignment and effective integration of heterogeneous sensor inputs within the language model.

\textbf{Low-Rank Adaptation (LoRA) in Stage \uppercase\expandafter{\romannumeral1}.}
In Stage I training, we utilize Low-Rank Adaptation (LoRA) to efficiently fine-tune the LLM backbone, minimizing the number of trainable parameters. Each LoRA module is characterized by a rank of 32, a scaling factor ($\alpha$) of 64, and a dropout rate of 0.05. LoRA modules do not include bias terms, further streamlining the adaptation process. This low-rank structure substantially reduces computational overhead, enabling stable and effective fine-tuning while maintaining the representational capacity required to integrate diverse multimodal information.

\textbf{Mixture of Expert in Stage \uppercase\expandafter{\romannumeral2}.}
In Stage \uppercase\expandafter{\romannumeral2}, we strategically integrate sparse Mixture-of-Experts (MoE) layers into selected transformer blocks ([8, 12, 16, 20, 24, 28]) within the language backbone. Each MoE layer consists of 8 parallel expert modules, implemented as specialized LLaMA-style multilayer perceptrons (MLP). These experts expand the feature dimension from 4096 to an intermediate dimension of 11008 via parallel gated projections (gate$\_$proj and up$\_$proj), followed by SiLU activations and dimensional reduction back to 4096 (down$\_$proj). Tokens are dynamically routed to these experts using a learnable Top-K gating network, which adaptively selects the most suitable experts based on token-level semantic characteristics. In our implementation, we set $K$ to 2. This design encourages sparsity and computational efficiency, enabling effective modeling of heterogeneous modality information while preserving scalability. The adaptive routing mechanism thus enhances the model’s ability to exploit specialized knowledge, significantly improving performance across diverse and complex multimodal 3D scene understanding tasks.

\end{document}